\documentclass{article}

\usepackage[eandd, preprint]{neurips_2026}

\makeatletter
\renewcommand{\@noticestring}{}
\makeatother

\usepackage[utf8]{inputenc}
\usepackage[T1]{fontenc}
\usepackage{hyperref}
\usepackage{url}
\usepackage{booktabs}
\usepackage{amsfonts}
\usepackage{amsmath}
\usepackage{amssymb}
\usepackage{nicefrac}
\usepackage{microtype}
\usepackage{xcolor}
\usepackage{graphicx}
\usepackage{subcaption}   % \captionof + side-by-side minipage floats (compact §4 layout)
\usepackage{xspace}
\usepackage{listings}
\usepackage{pifont}    % for \ding (checkmark / cross in tables)
\usepackage{tikz}
\usetikzlibrary{positioning, arrows.meta, shapes.geometric, fit, backgrounds}
\usepackage{pgfplots}
\pgfplotsset{compat=1.18}
\usepackage{siunitx}      % decimal-aligned numeric table columns (S[...])
\makeatletter
\newcommand{\atrexbench}{Atrex-Bench\xspace}
\newcommand{\atrexagent}{Atrex-Kernel-Agent\xspace}
\newcommand{\aka}{AKA\xspace}

\newcommand{\PHB}[1]{\noindent\textbf{#1}} % paragraph at the beginning
\newcommand{\PHM}[1]{\vspace{.4em} \noindent\textbf{#1}} % paragraph in the middle

\newcommand{\secref}[1]{\S\ref{#1}}
\newcommand{\figref}[1]{Figure~\ref{#1}}
\newcommand{\tabref}[1]{Table~\ref{#1}}

\newcommand{\eqnref}[1]{Equation~\ref{#1}}

\newcommand{\etc}{
        \@ifnextchar{.}
        \textit{etc}
        \textit{etc.\@\xspace}
}

\newcommand{\code}{\texttt}

\newcommand{\circledtext}[1]{\raisebox{.5pt}{\textcircled{\raisebox{-.9pt} {#1}}}}

\definecolor{mygreen}{rgb}{0,0.6,0}
\definecolor{mygray}{rgb}{0.5,0.5,0.5}
\definecolor{mymauve}{rgb}{0.58,0,0.82}
\definecolor{myred}{rgb}{0.79,0.15,0.15}
\lstdefinestyle{mypython}{
  language=Python,
  backgroundcolor=\color{white},   % choose the background color; you must add \usepackage{color} or \usepackage{xcolor}; should come as last argument
  basicstyle=\scriptsize\ttfamily,        % the size of the fonts that are used for the code
  breakatwhitespace=false,         % sets if automatic breaks should only happen at whitespace
  breaklines=true,                 % sets automatic line breaking
  captionpos=b,                    % sets the caption-position to bottom
  commentstyle=\color{mygreen},
  deletekeywords={...},            % if you want to delete keywords from the given language
  escapeinside={\%*}{*)},          % if you want to add LaTeX within your code
  extendedchars=true,              % lets you use non-ASCII characters; for 8-bits encodings only, does not work with UTF-8
  firstnumber=1,                % start line enumeration with line 1000
  frame=no,                    % adds a frame around the code
  keepspaces=true,                 % keeps spaces in text, useful for keeping indentation of code (possibly needs columns=flexible)
  keywordstyle=\color{mymauve},       % keyword style
  language={Python},                 % the language of the code
  morekeywords={DIRECT, PERIODIC, IMMEDIATE, BY_TIME, EVERY_OBJ, deferred},            % if you want to add more keywords to the set
  numbers=left,                    % where to put the line-numbers; possible values are (none, left, right)
  numbersep=5pt,                   % how far the line-numbers are from the code
  numberstyle=\tiny\color{mygray}, % the style that is used for the line-numbers
  rulecolor=\color{black},         % if not set, the frame-color may be changed on line-breaks within not-black text (e.g. comments (green here))
  showspaces=false,                % show spaces everywhere adding particular underscores; it overrides 'showstringspaces'
  showstringspaces=false,          % underline spaces within strings only
  showtabs=false,                  % show tabs within strings adding particular underscores
  stepnumber=1,                    % the step between two line-numbers. If it's 1, each line will be numbered
  stringstyle=\color{myred},     % string literal style
  tabsize=2,                    % sets default tabsize to 2 spaces
  title=\lstname                   % show the filename of files included with \lstinputlisting; also try caption instead of title
}

\makeatother

\title{Are LLM-Generated GPU Kernels Production-Ready? A Trace-Driven Benchmark and Optimization Agent}

\author{%
  \bfseries Lingyun Yang\thanks{Equal contributions.} \quad
  Yuxiao Wang\footnotemark[1] \quad
  Shenghao Liang \quad
  Linfeng Yang \\
  \bfseries Daocheng Ying \quad
  Chunbo You \quad
  Rui Zhang \quad
  Luping Wang \\
  \bfseries Yinghao Yu \quad
  Guodong Yang \quad
  Liping Zhang \\[0.5em]
  \normalfont ATREX Team\thanks{ATREX = Alibaba Tensor Research \& Engineering for XPUs.}, Alibaba Group
}

\begin{document}

\maketitle
\thispagestyle{plain}
% Author/title \thanks markers (e.g. equal-contribution *) advance the footnote
% counter on the NeurIPS title page; reset so body footnotes start at 1.
\setcounter{footnote}{0}

\begin{abstract}
Existing GPU kernel generation benchmarks draw problems from synthetic or curated sources that diverge from deployed workloads. We present \textbf{\atrexbench},\footnote{\url{https://github.com/alibaba/atrex-bench}} a benchmark whose 30 operators and 440 shapes are sampled directly from full-cluster production inference traces of compute-limited, memory-rich GPUs. Each problem carries an importance weight derived from its share of observed GPU time, weighted by application card-hours and computed separately for the serving phases in which it runs, together with a per-problem roofline ceiling, so the aggregate score emphasizes the kernels that consume the most serving time. Evaluating six frontier coding agents on \atrexbench shows that even the best vanilla model reaches only ${\sim}10\%$ of the hardware roofline on production operators---and correctness alone overstates capability, since much of the apparent pass rate comes from PyTorch fallbacks rather than kernels the model wrote. To close this gap, we co-release \textbf{\atrexagent} (\aka),\footnote{\url{https://github.com/alibaba/atrex-kernel-agent}} a profile-driven kernel-optimization agent that combines iterative measure--revise search, optimization dropout for escaping stalled search contexts, and a layered GPU-optimization knowledge base (298 reference-kernel files and 244 optimization-knowledge documents, plus external upstream reference projects for API/ISA lookup). In a controlled case study, the agent converts zero-FlyDSL fallbacks into real kernels that match or exceed hand-tuned production baselines.
\end{abstract}

\section{Introduction}
\label{sec:intro}

LLM coding agents can now write GPU kernels that compile, pass correctness checks, and approach hand-written performance on simple workloads. A line of benchmarks has tracked this progress, contributing the PyTorch-reference task format, roofline-style scoring against hardware ceilings, multi-vendor targets, and production hot-swapping~\citep{ouyang2025kernelbench, saroufim2025backendbench, li2025tritonbench, wen2025multikernelbench, xing2026flashinferbench, zhu2026cudabench, lin2026solexecbench}. The natural next question is whether these agents are ready for \emph{production}: can they write the kernels a real serving stack actually runs, on the hardware that fleet actually deploys, well enough to replace or improve production kernels?

Existing benchmarks cannot answer this, because their problems are synthetic or curated and so diverge from deployed workloads on three axes that decide production value. First, \emph{which shapes}: a production fleet runs a heavily skewed distribution---in our traces, the top five operators carry $\approx\!64\%$ of GPU wall-time---that a uniform synthetic grid does not reproduce. Second, \emph{which kernels matter}: an unweighted task average treats a rare elementwise op like a fused-attention path that dominates latency. Third, \emph{how good is good enough}: a kernel must approach the hardware roofline on its own shape, not merely beat an unoptimized baseline. A benchmark that misses these axes reports scores that do not predict deployability.

The workload axes force a different benchmark contract. The workload source must come from online serving traces rather than a hand-written operator list; the problem weights must preserve production skew rather than average every task equally; and the denominator must be a per-shape hardware ceiling rather than a mutable software baseline. The contract should also hide production provenance and roofline answers during generation; otherwise, an agent can exploit upstream names, known kernels, or the scoring formula instead of solving the kernel.

\textbf{\atrexbench} implements this contract. It samples problems from production inference traces on compute-limited, memory-rich GPU fleets spanning XPU-A\footnote{We use XPU-A as a desensitized name for the non-NVIDIA accelerator evaluated in this paper.} and H20, with more than 10k deployed accelerators; the current release contains 30 operators and 440 hot shapes drawn from 1{,}303 profiles and ${\sim}20$ deployed models across vLLM, SGLang, AITER, and RTP-LLM~\citep{kwon2023vllm,zheng2024sglang,amd2025aiter,tan2026rtpllm}. It weights each (op, shape) by its serving-time share, scores each against a per-problem roofline ceiling, and hides upstream provenance and roofline artifacts from candidate agents. Evaluating six frontier coding agents exposes a gap that prior benchmarks miss: the best candidate reaches only $10.7\%$ of the reference-derived hardware roofline ($S_{\text{agg}}=0.107$), and no agent matches the deployed production kernel (\secref{sec:models}). We also analyze the generated code after evaluation and identify a form of correctness reward hacking: models can pass by delegating to PyTorch fallbacks while writing little target-DSL code (e.g., Qwen3.7-Max achieves $84.8\%$ correctness with only $43.8\%$ FlyDSL adoption~\citep{flydsl2025github}), so correctness overstates target-DSL use (\secref{sec:models-illusion}).

The residual gap is concentrated in domain knowledge---roofline reasoning, hardware-specific instruction selection, and accumulated tactics---rather than raw coding ability. Motivated by this, we build \textbf{\atrexagent} (\aka), which pairs a profile-driven measure--revise workflow with \emph{optimization dropout}---a partial restart that escapes stalled search contexts---and a layered optimization-knowledge base. In a controlled study \aka mitigates both failure modes: it closes the target-DSL-dominance gap on a weaker model by converting $0\%$-FlyDSL fallbacks into near-$100\%$-FlyDSL kernels ($6.7\times$ over the fallback on \code{attention\_forward}), and it narrows the residual roofline gap on a stronger one ($S$ $0.28\to0.42$), with kernels that overtake the hand-tuned production baseline on both attention operators (\secref{sec:skills}).

\PHB{Contributions.} In summary, this work contributes: (1) \textbf{\atrexbench}, the first kernel-generation benchmark sourced from full-cluster production traces and scored with an importance-weighted, per-problem roofline metric~\citep{williams2009roofline} (\secref{sec:bench-design}); (2) a release contract that packages production-derived references, shape sets, hidden provenance, hidden roofline artifacts, and refreshable importance weights; (3) an evaluation of six frontier coding agents that quantifies how far they remain from deployability---a best $S_{\text{agg}}=0.107$---and uses post-hoc target-DSL-dominance analysis to expose a measurable correctness illusion (\secref{sec:models}); and (4) \textbf{\aka}, a profile-driven optimization agent that closes much of this gap, turning fallbacks into real kernels that overtake hand-tuned production baselines (\secref{sec:skills}). Both artifacts are released as open source.

\section{Related Work}
\label{sec:related}

\subsection{LLM Kernel Generation Benchmarks}
\label{sec:related-bench}

KernelBench~\citep{ouyang2025kernelbench} pioneered LLM kernel-generation evaluation with a 250-task, three-level difficulty suite drawn from PyTorch reference modules. BackendBench~\citep{saroufim2025backendbench}, from the Meta PyTorch team, reframes the problem as ``ship a correct \& fast backend for PyTorch'' and adds a hot-swap path for replacing ATen kernels in place. TritonBench~\citep{li2025tritonbench} targets Triton kernels specifically. MultiKernelBench~\citep{wen2025multikernelbench} extends KernelBench to 285 recategorized tasks and ports the evaluation harness to NVIDIA L20, Huawei Ascend NPU, and Google TPU. FlashInfer-Bench~\citep{xing2026flashinferbench} establishes a closed-loop framework (Trace schema $+$ leaderboard $+$ dynamic substitution via \code{apply()}) atop the FlashInfer engine, blurring the line between benchmarking and production hot-swapping. More recently, CUDABench~\citep{zhu2026cudabench} reframes the problem as \emph{text-to-CUDA} (rather than \emph{PyTorch-to-CUDA}) and introduces a roofline-based score over a 1{,}500-prompt set sampled from open-source CUDA repositories. NVIDIA's SOL-ExecBench~\citep{lin2026solexecbench} targets ``Speed-of-Light''-style hardware-utilization measurement.

\atrexbench shares the goal of measuring LLM kernel generation and reuses several conventions established above (PyTorch-defined references, three-stage compile/correctness/perf gating, roofline-style scoring); it adds two complementary capabilities visible at a glance in \tabref{tab:bench-matrix}: it draws its problem set from \emph{online production traces} rather than synthetic or curated tasks, and it scores kernels under an \emph{importance-weighted} aggregate that reflects production time share. We also plan to refresh the problem distribution from production traffic as workloads evolve, so the benchmark can track deployed needs while preserving versioned snapshots for reproducibility.

\begin{table}
    \caption{Cross-benchmark capability matrix (positioning, not ranking), ordered by release. ``Roofline'': ships a per-problem roofline/Speed-of-Light bound; ``Imp.-wtd.'': importance-weighted aggregate; ``Prod.-sampled'': problems sampled from production traffic.}
    \label{tab:bench-matrix}
    \centering
    \small
    \begin{tabular}{lllll}
        \toprule
        Benchmark & Source & Roofline & Imp.-wtd. & Prod.-sampled \\
        \midrule
        KernelBench~\citep{ouyang2025kernelbench}        & synth.\ PyTorch         & \ding{55} & \ding{55} & \ding{55} \\
        BackendBench~\citep{saroufim2025backendbench}    & PyTorch ATen + traces   & \ding{55} & \ding{55} & partial \\
        TritonBench~\citep{li2025tritonbench}            & hand-crafted            & \ding{55} & \ding{55} & \ding{55} \\
        MultiKernelBench~\citep{wen2025multikernelbench} & KernelBench fork        & \ding{55} & \ding{55} & \ding{55} \\
        FlashInfer-Bench~\citep{xing2026flashinferbench} & curated + traces        & \ding{55} & \ding{55} & partial \\
        CUDABench~\citep{zhu2026cudabench}               & public CUDA repos       & \ding{51} & \ding{55} & \ding{55} \\
        SOL-ExecBench~\citep{lin2026solexecbench}        & curated                 & \ding{51} & \ding{55} & \ding{55} \\
        \textbf{\atrexbench}                              & \textbf{online production trace} & \textbf{\ding{51}} & \textbf{\ding{51}} & \textbf{\ding{51}} \\
        \bottomrule
    \end{tabular}
\end{table}

\subsection{Roofline Model and Hardware Performance Bounds}
\label{sec:related-roofline}

The roofline model~\citep{williams2009roofline} bounds a kernel's achievable performance by $\min(\text{P}_{\text{peak}},\;\text{AI}\cdot\text{B}_{\text{peak}})$, giving a hardware-independent utilization metric widely used in GPU performance engineering. CUDABench~\citep{zhu2026cudabench} and SOL-ExecBench~\citep{lin2026solexecbench} are the closest prior benchmarks to adopt this metric; \atrexbench differs in deriving a versioned roofline ceiling for each production-derived problem and feeding per-kernel roofline achievement into an importance-weighted aggregate rather than a uniform mean.

\subsection{Skills and Knowledge Libraries for Code Agents}
\label{sec:related-skills}

Equipping an agent with a retrievable skill library has been explored in embodied settings (Voyager~\citep{wang2023voyager}), general reasoning (Self-Discover~\citep{zhou2024selfdiscover}), and software engineering (SWE-agent~\citep{yang2024sweagent}; Anthropic Skills~\citep{anthropic2025skills}). Recent GPU-kernel systems bring the same idea into optimization: AKO~\citep{ako2026} packages existing coding agents in an optimization harness with benchmark/profiler interfaces and closed-loop campaigns, while KDA~\citep{kda2026github} codifies a CUDA-kernel workflow around task contracts, planning, verification, profiles, and reusable kernel knowledge. \aka differs in two main ways. First, its \emph{optimization dropout} mechanism masks stale iteration memories while preserving the accepted kernel and audit trail, giving a fresh sub-agent enough context to explore a different optimization direction rather than remaining trapped in a local optimum. Second, it uses a layered knowledge base that combines expert-contributed optimization experience and hardware facts with reference kernels and best practices retrieved from upstream open-source projects. Together with official-profiler feedback, these mechanisms turn knowledge retrieval into an iterative search process rather than a one-shot injection of additional context (\secref{sec:skills-main}).

\section{\atrexbench: Design}
\label{sec:bench-design}

\begin{figure}[t]
    \centering
    \includegraphics[width=\linewidth]{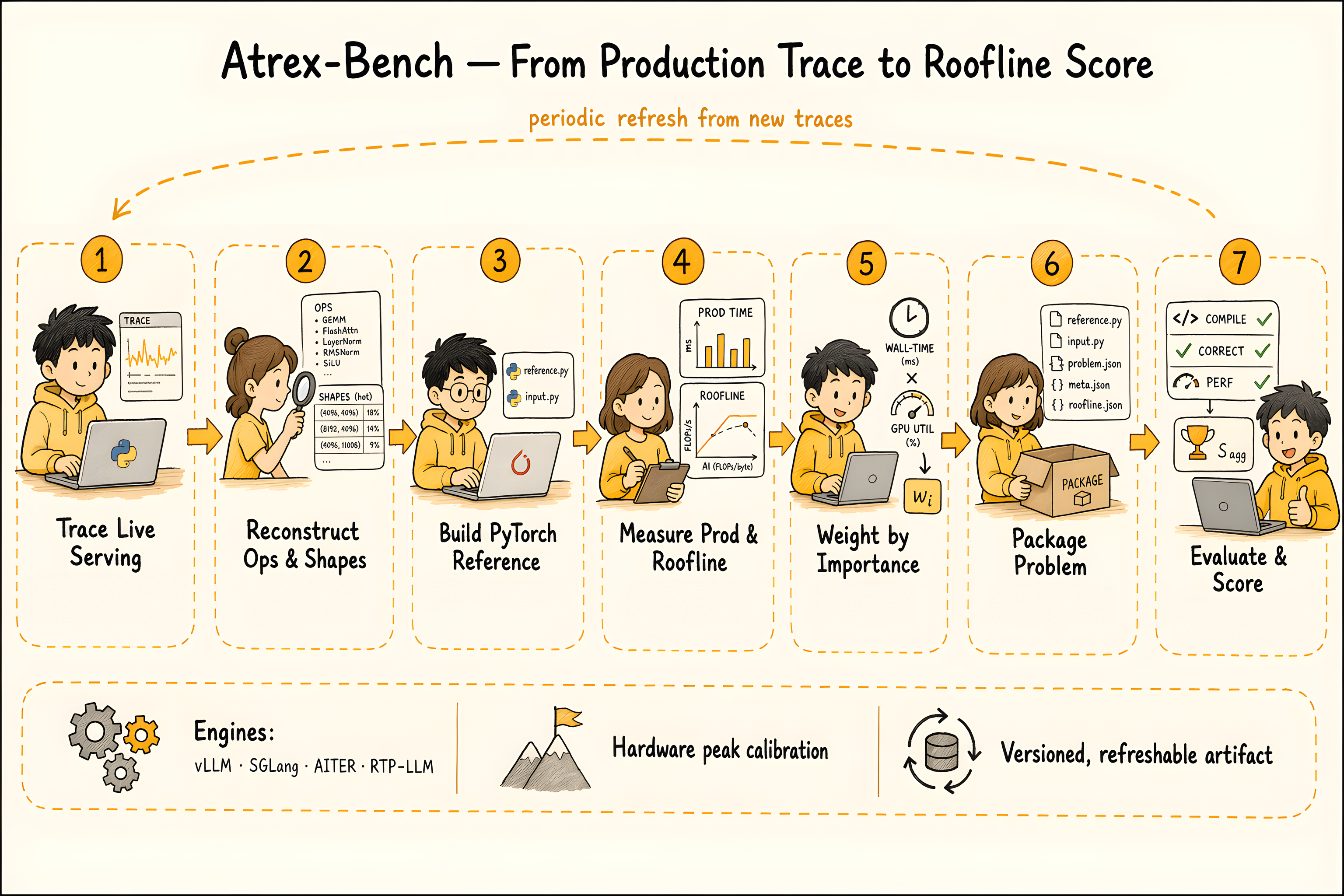}
    \caption{\atrexbench lifecycle, from production trace to roofline score: trace
    live serving, reconstruct operators and shapes, build PyTorch references,
    derive per-problem roofline bounds and importance weights, package the
    hidden-provenance problem set, and run the three-stage
    compile/correctness/performance gate.}
    \label{fig:bench-pipeline}
\end{figure}

This section describes how \atrexbench constructs benchmark operators and shape-level artifacts from production traces (Sections~\ref{sec:bench-source}--\ref{sec:bench-spec}), how it derives per-problem roofline bounds and importance weights (Sections~\ref{sec:bench-roofline}--\ref{sec:bench-weighting}), how it scores candidate kernels (\secref{sec:bench-score}), and how it evaluates submitted kernels (\secref{sec:bench-procedure}). Figure~\ref{fig:bench-pipeline} provides an end-to-end overview of this lifecycle.

\subsection{From Production Traces to Operators and Shapes}
\label{sec:bench-source}
\label{sec:bench-pipeline}

The benchmark's problem set is sourced from online production traces collected
on selected XPU-A and H20 inference clusters with more than 10k deployed
accelerators. The tracer attaches to running workloads without relaunching the
service or inserting application-level instrumentation. It combines Python-frame
context (operator phase, batch/token shape, framework state) with vendor runtime
traces (kernel launches, timing, streams, memory operations, and graph anchors),
then correlates host and device events through correlation IDs when available and
timestamps or graph anchors otherwise. This gives a cross-layer view of which
framework operation produced each device kernel and under which serving shape.

\PHB{From traces to benchmark units.} The construction pipeline maps raw kernel
records into computation-level operator families rather than framework-specific
kernel names, so semantically identical operators from vLLM~\citep{kwon2023vllm},
SGLang~\citep{zheng2024sglang}, AITER~\citep{amd2025aiter}, and
RTP-LLM~\citep{tan2026rtpllm} can share one benchmark problem. Each released
(operator, shape) entry retains upstream provenance in \code{metadata.json} for
reproducibility.

\PHM{Reference and shape construction.} For each operator, we reimplement the
production logic as a hardware-agnostic PyTorch reference
(\code{reference.py}) and validate it against the production kernel. Shape sets
come directly from traced tensor dimensions when available; otherwise, fixed
model dimensions are recovered from deployment configuration and dynamic
dimensions (e.g., token count, batch size) are set to deployment-typical
values. Because raw traces contain many near-duplicate variants, the pipeline
keeps architecture-fixed dimensions exact and buckets dynamic token dimensions
from single-token decode to long-context prefill. The current release contains
\textbf{30} operators and \textbf{440} representative shapes.

\PHM{Baselines and weights.} For each retained (operator, shape), we record the
production framework kernel's observed production runtime as a deployment baseline and a
sanity check for the roofline bound. The pipeline also assigns each operator a
phase label (\code{prefill}, \code{decode}, or \code{prefill/decode}) and derives
its importance weight $w_i$ from its share of observed device time, weighted by
application card-hours and computed separately for the serving phases in which it runs
(\secref{sec:bench-weighting}).

\subsection{Problem Specification Format}
\label{sec:bench-spec}

Each \atrexbench operator is shipped as a directory containing a fixed five-file
contract, summarized in \tabref{tab:problem-spec}.

\begin{table}[t]
    \caption{Per-operator problem specification files in \atrexbench; ``Visible'' marks files available to candidate agents during generation.}
    \label{tab:problem-spec}
    \centering
    \small
    \begin{tabular}{@{}p{0.18\linewidth}p{0.67\linewidth}c@{}}
        \toprule
        File & Contents & Visible \\
        \midrule
        \code{reference.py} &
        Executable PyTorch specification of the operator semantics. &
        \ding{51} \\
        \code{input.py} &
        Input generator for randomized and corner-case tensors. &
        \ding{51} \\
        \code{shapes.json} &
        Production-derived shape configurations, including model dimensions and serving-time dynamic dimensions. &
        \ding{51} \\
        \code{metadata.json} &
        Release metadata: operator identity, dtype information, production baseline time, and upstream provenance. &
        \ding{55} \\
        \code{roofline.json} &
        Scoring metadata: semantic work/traffic estimates and per-hardware speed-of-light time. &
        \ding{55} \\
        \bottomrule
    \end{tabular}
\end{table}

\PHB{Generation--evaluation boundary.} The five-file contract separates the
operator specification used for generation from the artifacts used only by the
evaluator. During generation, the visible surface is the executable problem
statement: \code{reference.py}, \code{input.py}, \code{shapes.json}, and the
prompt constraints.
For generation CLIs with explicit tool controls, we further restrict the callable
tool surface: default runs expose local workspace inspection, search, editing,
and shell-based build/test iteration, while web-retrieval, notebook-editing, and
delegation and auxiliary-agent tools are excluded from comparable benchmark runs. Optimizer-augmented
experiments that intentionally enable these tools are therefore treated as a
separate setting.
The hidden files provide provenance and scoring state to the evaluator, but are
not part of the generation prompt: \code{metadata.json} contains upstream
provenance, and \code{roofline.json} contains the per-shape scoring denominators.
During evaluation, the generated file is imported into a fresh environment that
contains the full benchmark state and evaluator, decoupling what the model can
read from how its output is judged. This boundary also prevents direct leakage
from answer-bearing artifacts, such as recovering the source kernel by name or
tuning directly to the roofline denominator.
Per-operator importance weights (\secref{sec:bench-weighting}) are shipped as a
release-level artifact, so that re-weighting on a new trace does not require
touching individual problem assets.

\subsection{Per-Problem Roofline Bound Derivation}
\label{sec:bench-roofline}

The roofline bound gives a hardware lower bound on the latency of each benchmark
unit. We derive it in two steps. First, accelerator compute throughput and memory
bandwidth peaks are calibrated once and recorded as release constants. Second,
the reference implementation defines the unit's semantic work $F_j$ and memory
traffic $M_j$. For a target accelerator with dtype-specific compute peak
$P_{\tau_j}$ and memory bandwidth $\beta$, the speed-of-light latency for unit
$j$ is
\begin{equation}
T_{\text{roofline},j}
= \max\!\left(\frac{F_j}{P_{\tau_j}},\, \frac{M_j}{\beta}\right).
\label{eq:roofline-time}
\end{equation}
This $T_{\text{roofline},j}$ is stored in \code{roofline.json} and serves as the
per-problem denominator in the roofline achievement score
(\secref{sec:bench-score}).

\subsection{Importance Weighting}
\label{sec:bench-weighting}

Each operator is assigned an importance weight.

\PHB{Importance weight $w_i$.} The weight estimates an operator's share of
production GPU time while preserving both the deployed application mix and the
prefill/decode distinction. Let $a$ index applications and let $p_a$ be
application $a$'s fraction of fleet card-hours, with $\sum_a p_a=1$. Operator
$i$ may comprise multiple device kernels. If kernel $r$ is invoked
$m_{a,\phi,r}$ times in application $a$ during phase
$\phi\in\{\text{prefill},\text{decode}\}$, and invocation $\ell$ has duration
$d_{a,\phi,r,\ell}$, its cumulative observed device time is
\begin{equation}
D_{i,a}^{\phi}
= \sum_{r\in\mathcal{K}_i}\sum_{\ell=1}^{m_{a,\phi,r}}
d_{a,\phi,r,\ell},
\qquad
D_a^{\phi}=\sum_j D_{j,a}^{\phi}.
\label{eq:imp-time}
\end{equation}
The ratio $D_{i,a}^{\phi}/D_a^{\phi}$ is therefore the fraction of observed GPU
time consumed by operator $i$ within that application's collection window for
phase $\phi$; invocation frequency is represented by the sum over $m_{a,\phi,r}$.
Let $\Phi_i$ denote the phases in which operator $i$ runs. We use the single
phase share for a prefill-only or decode-only operator and the simple average of
the two phase shares for an operator that runs in both phases. Weighting this
quantity by the application's fleet card-hour share gives
\begin{equation}
\tilde w_i
= \sum_a p_a\,
  \frac{1}{|\Phi_i|}
  \sum_{\phi\in\Phi_i}\frac{D_{i,a}^{\phi}}{D_a^{\phi}},
\qquad
w_i=\frac{\tilde w_i}{\sum_j\tilde w_j},
\qquad \textstyle\sum_i w_i=1 .
\label{eq:imp-weight}
\end{equation}
The final normalization restricts the distribution to the operators retained in
the benchmark release.
The current release aggregates \textbf{1{,}303} production profiles spanning four
serving frameworks (vLLM, SGLang, AITER, and RTP-LLM), and the weight distribution is heavily skewed: the top five
operators carry $\approx\!\textbf{64\%}$ of the weight (\code{unified\_attention}
$36.1\%$, \code{fused\_moe} $10.4\%$, \code{block\_scaled\_mm} $8.5\%$,
\code{fp8\_blockscale\_fused\_moe} $4.7\%$, \code{paged\_attention\_decode}
$4.0\%$), while the remaining $25$ operators share the rest. These weights are
recomputed from production traces whenever the benchmark distribution is refreshed.

\eqnref{eq:imp-weight} is what makes the aggregate \emph{production-weighted}
rather than suite-uniform: an operator that dominates deployed wall-time dominates
the score, however many or few shapes it contributes.

\subsection{Operator and Shape Distribution}
\label{sec:bench-distribution}

The current release covers \textbf{30} production operators with \textbf{440} hot shapes in total. Several structural properties of the resulting distribution are worth surfacing because they shape what the benchmark actually measures.

\PHB{Heavy-tailed importance.} Importance weights are concentrated in a small number of operators rather than spread evenly. \tabref{tab:top-ops} reports the ten operators with the largest app-card-hour-weighted, phase-aware GPU-time shares: the top five (\code{unified\_attention}, \code{fused\_moe}, \code{block\_scaled\_mm}, \code{fp8\_blockscale\_fused\_moe}, \code{paged\_attention\_decode}) together account for roughly $\mathbf{64\%}$ of the normalized production importance, and the top ten for roughly $\mathbf{80\%}$. The remaining 20 operators each contribute under $3\%$ but still cover several distinct families (RoPE, RMS-norm variants, MoE-gating utilities, paged-cache management, fp8 dynamic quantization, top-$k$ filtering). The aggregate score in \secref{sec:bench-score} therefore rewards making the head right while still requiring correctness on the long tail.

\begin{table}[t]
    \caption{Top-10 \atrexbench operators by importance weight $w_i$ (\eqnref{eq:imp-weight}). ``Score'' is the normalized app-card-hour-weighted, phase-aware GPU-time share; ``Shapes'' the number of shipped (init/input) configurations; ``Dtype'' the operator's primary precision.}
    \label{tab:top-ops}
    \centering
    \small
    \begin{tabular}{lcccl}
        \toprule
        Operator & Score $w_i$ & Cum.\ & Shapes & Dtype \\
        \midrule
        \code{unified\_attention}         & 0.361 & 0.361 & 25 & bf16 \\
        \code{fused\_moe}                 & 0.104 & 0.465 & 23 & bf16 \\
        \code{block\_scaled\_mm}          & 0.085 & 0.550 & 24 & fp8\_e4m3 \\
        \code{fp8\_blockscale\_fused\_moe}& 0.047 & 0.596 &  6 & fp8\_e4m3 \\
        \code{paged\_attention\_decode}   & 0.040 & 0.637 &  8 & bf16 \\
        \code{reshape\_and\_cache}        & 0.040 & 0.677 &  8 & bf16 \\
        \code{topk\_filter}               & 0.032 & 0.709 &  8 & fp32 \\
        \code{gated\_delta\_rule\_update} & 0.032 & 0.741 & 17 & bf16 \\
        \code{fused\_qkv\_rope}           & 0.031 & 0.771 &  6 & fp16 \\
        \code{rms\_norm}                  & 0.025 & 0.796 & 56 & bf16 \\
        \midrule
        20 tail operators (combined)        & 0.204 & 1.000 & 259 & mixed \\
        \bottomrule
    \end{tabular}
\end{table}

\PHM{Mixed precision is the norm.} The 30 operators span five distinct precisions---bf16 (19 operators), fp8\_e4m3 (5), fp16 (2), fp32 (2), and int32 (2). Operators that look superficially similar can ship at different precisions: \code{fused\_moe} runs in bf16 while \code{fp8\_blockscale\_fused\_moe} runs in fp8\_e4m3, and \code{block\_scaled\_mm} uses fp8\_e4m3 throughout. Two consequences flow from this for the methodology: hardware peaks $P_{\text{peak}}$ must be parameterized per-dtype (\secref{sec:bench-roofline}), and a candidate kernel cannot achieve a high roofline score by silently up-casting to a more peak-favorable dtype, because the correctness gate (\secref{sec:bench-procedure}) compares against the reference at its declared dtype.

\PHM{Shape count is decoupled from importance.} The number of shapes per operator (column ``Shapes'' in \tabref{tab:top-ops}) reflects the operator's \emph{shape diversity} in production, not its importance. \code{rms\_norm} ships 56 shapes but carries only $2.5\%$ of the weight; \code{paged\_attention\_decode} ships 8 shapes but carries $4.0\%$. This decoupling is why the aggregate is weighted by $w_i$ rather than by raw $(\text{op}, \text{shape})$ count: a candidate that aces 50 rms-norm variants and fails attention should not outrank one that does the reverse.

\PHM{Compute- vs.\ memory-bound regime.} The (operator, shape) pairs span a wide range of arithmetic intensity (AI $=$ FLOPs / bytes moved), and the regime depends on both the operator and the shape: a GEMM-based operator like \code{fused\_moe} is memory-bound at single-token decode but compute-bound at large-batch prefill. Across the 440 shapes, the compute-bound end is dominated by large-batch GEMM and attention shapes (AI $> 100$), while normalization and activation shapes (\code{rms\_norm}, \code{silu\_and\_mul}; AI $< 1$) are bandwidth-limited regardless of token count, and pure data-movement operators (\code{reshape\_and\_cache}, \code{topk\_filter}) perform no arithmetic at all. The absolute scale varies accordingly: a single large-batch \code{fp8\_blockscale\_fused\_moe} shape performs ${\sim}400$B FLOPs over ${\sim}7$\,GB, while a single-token \code{fused\_qkv\_rope} shape performs ${\sim}15$K FLOPs over ${\sim}29$\,KB---seven orders of magnitude in both dimensions. The benchmark thus exercises both regimes across representative token lengths---a property that proves diagnostic in \secref{sec:models-regime}, where agents reach several times more of the roof on memory-bound shapes than on compute-bound ones.

\PHM{Production model and provenance coverage.} The 440 shapes are drawn from ${\sim}20$ production-model deployments spanning MoE architectures (e.g., Qwen3 MoE variants and DeepSeek-R1), vision--language models (Qwen3-VL), and dense models (QwQ-32B, Qwen3-32B, and Qwen3-8B). Because the open-source metadata redacts exact model identities, the public corpus exposes this diversity through operator provenance: assigning each mixed-origin operator to its primary upstream framework, the 440 shapes break down into vLLM (292), SGLang (90), RTP-LLM (38), and AITER (20). The same metadata also records the production baseline backend used for XPU-A measurement, which is AITER for 266 shapes, SGLang for 151, and RTP-LLM for 23. This skew reflects the deployment mix on the traced cluster rather than an artificial balance; the benchmark therefore captures a shared operator vocabulary across a heterogeneous production fleet, not a single model's kernel mix.

\subsection{Roofline Score}
\label{sec:bench-score}

A candidate is scored at three levels---per shape, per operator, and as a single
production-weighted aggregate---so that the headline number reflects
kernel-generation ability \emph{as it would be felt in the serving environment},
not average competence over a suite where every operator counts equally.

\PHB{Per-shape achievement.} For an evaluation unit $j=(i,s)$ (operator $i$,
shape $s$) that passes correctness, with measured candidate time $T_{\text{cand},j}$
and per-shape roofline (speed-of-light) time $T_{\text{roofline},j}$
(\secref{sec:bench-roofline}),
\begin{equation}
S_j = \frac{T_{\text{roofline},j}}{T_{\text{cand},j}} \in (0,1].
\label{eq:per-kernel}
\end{equation}
$T_{\text{roofline}}$ is a hardware lower bound, so a value above $1$ indicates
inconsistent units, measurement error, or an invalid bound and is treated as an
evaluation error rather than clipped. $T_{\text{roofline},j}$ is derived from the
\emph{reference} semantics, never from the candidate's own profile, and is fixed
across candidates. $S_j$ is left undefined when the unit fails to compile or to
match the reference.

\PHM{Per-operator achievement.} Because shape counts vary widely across
operators ($3$ to $56$ in the current release), we summarize each operator by the
median over its correct shapes, and assign zero where the candidate produces no
correct kernel at all:
\begin{equation}
S_i =
\begin{cases}
\operatorname{median}\{\,S_j : j=(i,s)\ \text{correct}\,\}, & \text{operator $i$ has a correct shape,}\\[2pt]
0, & \text{otherwise.}
\end{cases}
\label{eq:per-op}
\end{equation}
The zero is deliberate: an operator for which no correct kernel is generated
delivers no production value and must not be silently dropped from the aggregate.

\PHM{Importance-weighted roofline score.} The headline metric weights each
operator's achievement by its production importance $w_i$ (\eqnref{eq:imp-weight}):
\begin{equation}
S_{\text{agg}} = \sum_i w_i\,S_i \;\in[0,1].
\label{eq:agg}
\end{equation}
$S_{\text{agg}}$ is the production-importance-weighted fraction of the hardware
roofline the candidate attains. It is high only when a candidate generates fast,
correct kernels for the operators that dominate deployed wall-time: acing a rare
operator barely moves it, while failing a heavy one---which enters as $S_i=0$---costs
its full weight. This is what lets $S_{\text{agg}}$ measure real-environment
kernel-generation ability. We use the weighted arithmetic mean rather than a
geometric mean, since the latter collapses to $0$ on a single failed operator
($S_i=0$) and erases all other signal. Operationally, the evaluator first computes
the per-shape score $S_j$ for each correct unit, summarizes each operator by
the median of its correct shapes (or $0$ if none pass), and then applies the
production weights in \eqnref{eq:agg}.

\subsection{Evaluation Procedure}
\label{sec:bench-procedure}

Beyond this separation boundary, each submitted kernel is evaluated through a three-stage gate: (1) \emph{compile}, (2) \emph{correctness}---compared against \code{reference.py} on randomized and corner-case inputs using the evaluator's numerical-equivalence check, and (3) \emph{performance}---measured in a controlled sandbox with an explicit frequency-lock check, per-iteration L2 flush, and per-shape subprocess isolation. The evaluator returns the candidate's per-shape wall-clock latency $T_{\text{cand},j}$. Only kernels that reach stage (3) contribute to $S_j$.

\medskip
\noindent
With the benchmark design in place, \secref{sec:models} turns to the empirical question that motivates the rest of the paper: how do current LLM agents perform on \atrexbench, and where do they fall short?

\section{Evaluating Frontier Agents on \atrexbench}
\label{sec:models}

We run \atrexbench on a panel of frontier coding agents, both to report where
today's models stand and to test whether the benchmark separates systems that
coarser measures rank as equal. Relative-speedup benchmarks
lose this resolving power on production serving stacks: once a model learns to
call an optimized library, it can pass without writing a kernel at all. We
therefore measure at the granularity of a single (op, shape) unit and ask where
each generated kernel stops behaving like a deployable one---at compilation, at
numerical correctness, at target-DSL dominance, or at the hardware roof.

\subsection{Setup}
\label{sec:models-setup}

\PHB{Operators and shapes.} The current \atrexbench release covers
\textbf{30} production operators, with \textbf{440} hot shapes in
total; one full evaluation pass therefore covers 440 (op, shape) pairs.
Correctness is checked against the eager reference over $K=5$ random seeds with
tolerances $\text{atol}=10^{-2}$, $\text{rtol}=5\!\times\!10^{-2}$; a shape
counts as correct only when all five seeds pass. Performance is the median
end-to-end forward time after warmup. Operators carry very different shape counts
(from $3$ to $56$ each), so every aggregate below is \emph{operator-balanced}:
each operator is first summarized over its own shapes and then weighted equally,
so a high-shape-count operator does not dominate the score.

\PHM{Hardware.} The experiments in this section are conducted primarily on XPU-A.

\PHM{Target DSL and agent runtime.} We deliberately target \textbf{FlyDSL}, a
DSL essentially absent from the models' pre-training corpora, so the benchmark tests
\emph{learning} rather than recall: a candidate must acquire an unfamiliar programming
model from the provided references and follow its constraints precisely, jointly probing
in-context learning, instruction-following, and code generation. (The prompt framework
also supports Triton, Gluon, and CuteDSL; we leave those to future releases.) Every
candidate receives the same prompt and tool set, and the runtime is held fixed---Claude
Code, except GPT-5.5 on its native Codex runtime---so the comparison reflects base-model
capability rather than harness differences.

\PHM{Candidates.} We evaluate six frontier coding agents---Claude
Opus~4.7, GPT-5.5, Qwen3.7-Max, Kimi-K2.6, GLM-5.1, and DeepSeek-V4-Pro---on
the full set of 30 operators and 440 units.

\PHM{Reference baselines.} Two non-LLM reference points anchor the ``vs''
columns. The first is \code{torch.compile} applied to the reference
implementation. The second is the deployed production kernel for each operator
(AITER / vLLM / SGLang / RTP-LLM), whose per-shape time is recorded in
\code{metadata.json}. These are reference baselines, not candidates, and appear
only in the ``vs'' columns of \tabref{tab:main-xpu-a}.

\PHM{Metrics.} Each metric first summarizes an operator over its shapes,
then aggregates across the $30$ operators. Bounded rates use the operator
\emph{mean}; the heavy-tailed time ratios use the operator \emph{median}, so a
single outlier operator cannot dominate the aggregate.
\begin{itemize}
    \item \emph{Compile rate}: the mean across operators of (compiled shapes $/$ shapes). A shape compiles only when it produces a completed ahead-of-time MLIR artifact, not merely a module that imports.
    \item \emph{Correctness}: the mean across operators of (correct shapes $/$ shapes), where a shape is correct iff it compiles \emph{and} matches the eager reference on all $K{=}5$ seeds; compile failures therefore count as incorrect.
    \item \emph{FlyDSL adoption}: the mean across operators of each operator's mean per-shape FlyDSL device-time share---the fraction of forward-pass device time spent inside \code{@flydsl.kernel}---which distinguishes target-DSL-dominant execution from paths that delegate primarily to PyTorch or precompiled vendor operators.
    \item \emph{Roofline achievement} $S_{\text{agg}}$: per shape, $S_j=T_{\text{SOL},j}/T_{\text{cand},j}$, with $T_{\text{SOL}}$ derived from the \emph{reference} semantics (\secref{sec:bench-roofline}) and fixed across candidates; values above $1$ indicate an evaluation error rather than being clipped. Per operator, $S_i$ is the median over its correct shapes ($0$ if none). The reported headline is the \emph{importance-weighted} aggregate $S_{\text{agg}}=\sum_i w_i S_i$ (\secref{sec:bench-score}), weighting each operator by its production wall-time share $w_i$ so the score reflects ability where production time actually goes.
    \item \emph{Reference ratios} $T_{\text{torch.compile}}/T_{\text{cand}}$ and $T_{\text{prod}}/T_{\text{cand}}$, reported as the operator median of per-operator medians over correct shapes (not importance-weighted); a value ${>}1$ means the candidate is faster than the baseline.
\end{itemize}
Compile rate and correctness average over all $30$ operators, so an operator a
model fails entirely contributes $0$. FlyDSL adoption, $S$, and the ratios are
taken only over the operators where they are defined (an operator with no
running or no correct shape is excluded, not scored $0$); $S$ and the ratios are
thus conditioned on each model's own correct operators and read as within-model
summaries. The failure-penalizing, production-weighted counterpart is the
importance-weighted aggregate $S_{\text{agg}}$ of \secref{sec:bench-score}.

\subsection{Main Results}
\label{sec:models-main}

\PHB{The evaluation gates separate the field before performance is considered.}
Compile rate ranges from $60.9\%$ to $100\%$, while correctness and target-DSL
adoption widen the separation further (\tabref{tab:main-xpu-a});
\secref{sec:models-errors} analyzes where these failures arise.

\PHM{Correctness and FlyDSL adoption carry most of the remaining separation.}
Correctness spans roughly forty-five points, from Opus~4.7 at $92.0\%$ down to GLM-5.1
at $46.2\%$, and FlyDSL adoption splits the field again: Opus~4.7 and GPT-5.5
spend $78.5\%$ and $71.6\%$ of their device time in FlyDSL, but the other four sit
below $44\%$---Qwen3.7-Max, for instance, reaches $84.8\%$ correctness with only
$43.8\%$ FlyDSL adoption, a gap we examine in \secref{sec:models-illusion}.

\PHB{The deployment-facing score reorders the leaders.} GPT-5.5 leads on both roofline
aggregations, and the gap widens under importance weighting. By the unweighted
operator median it edges Opus~4.7 ($0.129$ vs.\ $0.104$); the importance-weighted
$S_{\text{agg}}$, which credits each operator in proportion to its production
wall-time share (\secref{sec:bench-score}), puts it well ahead
($0.107$ vs.\ $0.059$, roughly $1.8\times$). The widening localizes Opus's
weakness: it reaches the roof on the typical, mostly bandwidth-bound operator but
falls far short on the production-heavy compute-bound operators that
$S_{\text{agg}}$ up-weights, where its median achievement is $0.009$ against
GPT-5.5's $0.074$ (\secref{sec:models-regime}). On the production-weighted
score, then, GPT-5.5 is the most hardware-efficient, while Opus~4.7's marginal
edge in raw correctness ($92.0\%$) does not carry to the roof where production
time concentrates. The
absolute numbers stay low: only Opus~4.7 and GPT-5.5 beat \code{torch.compile}
on the median operator ($2.29\times$ and $3.06\times$; Qwen3.7-Max is marginal at
$1.10\times$), and \emph{no} candidate beats the deployed production kernel---Opus~4.7
comes closest at $0.99\times$, GPT-5.5 reaches $0.85\times$, and the rest trail to
    $0.12\times$. Even the best candidate reaches only $10.7\%$ of the
    reference-derived hardware roofline ($S_{\text{agg}}=0.107$): passing correctness on production operators does
not imply production performance.

\begin{table}
    \caption{Main results on XPU-A; all
    metrics are operator-balanced over the $30$ operators. $S_{\text{agg}}$ is the
    importance-weighted roofline achievement (\secref{sec:bench-score}); the ``vs''
    columns are per-operator medians against \code{torch.compile} (\code{t.c.}) and
    the production kernel.}
    \label{tab:main-xpu-a}
    \centering
    \small
    \setlength{\tabcolsep}{5pt}
    \renewcommand{\arraystretch}{1.15}
    \begin{tabular}{l *{3}{S[table-format=3.1]} S[table-format=1.3] *{2}{S[table-format=1.2]}}
        \toprule
        & \multicolumn{3}{c}{Pass rate (\%)} & {Roofline} & \multicolumn{2}{c}{vs.\ baseline ($\times$)} \\
        \cmidrule(lr){2-4} \cmidrule(lr){6-7}
        {Model} & {Compile} & {Correct} & {FlyDSL} & {$S_{\text{agg}}$} & {\code{t.c.}} & {prod} \\
        \midrule
        Claude Opus 4.7 & 99.6 & \bfseries 92.0 & \bfseries 78.5 & 0.059 & 2.29 & \bfseries 0.99 \\
        GPT-5.5         & \bfseries 100.0 & 91.1 & 71.6 & \bfseries 0.107 & \bfseries 3.06 & 0.85 \\
        Qwen3.7-Max     & 97.1 & 84.8 & 43.8 & 0.047 & 1.10 & 0.19 \\
        Kimi-K2.6       & 91.5 & 81.5 & 40.1 & 0.043 & 0.94 & 0.33 \\
        GLM-5.1         & 60.9 & 46.2 & 38.6 & 0.015 & 0.97 & 0.33 \\
        DeepSeek-V4-Pro & 81.0 & 62.3 & 36.4 & 0.012 & 0.63 & 0.12 \\
        \bottomrule
    \end{tabular}
\end{table}

\subsection{The Correctness Illusion}
\label{sec:models-illusion}

\begin{figure}[t]
    \centering
    \includegraphics[width=\linewidth]{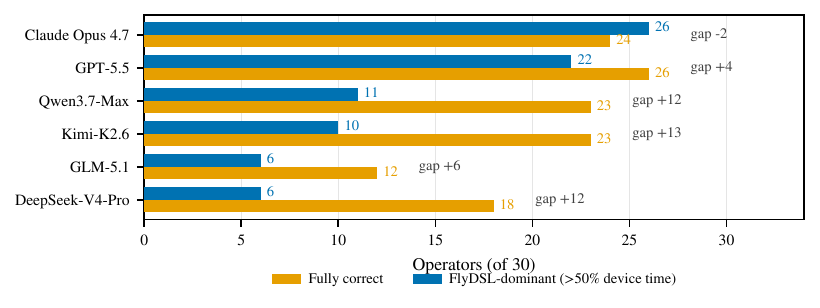}
    \caption{Correctness versus target-DSL dominance per candidate: operators that are fully
    correct (orange) versus those whose device time is FlyDSL-dominant (blue,
    $>\!50\%$); the gap is correctness carried by non-DSL fallbacks.}
    \label{fig:illusion}
\end{figure}

A passing kernel is not necessarily a written kernel. Because the compile gate
admits PyTorch fallbacks and calls into precompiled vendor kernels, a model can
accumulate correctness while producing little FlyDSL, and correctness alone then
overstates target-DSL use. Counting, per model, the operators that
are fully correct against those whose device time is genuinely dominated
($>\!50\%$) by FlyDSL exposes the gap (\figref{fig:illusion}). It is widest for
the middle of the field: Kimi-K2.6 ($23$ correct, $10$ FlyDSL-dominant),
Qwen3.7-Max ($23$ and $11$), and DeepSeek-V4-Pro ($18$ and $6$) all carry roughly
half their correct operators on non-DSL paths---answering attention with
\code{scaled\_dot\_product\_attention} and GEMM with a precompiled AITER entry
point. Opus~4.7 is the exception, with a \emph{negative} gap
($24$ correct, $26$ FlyDSL-dominant): it writes FlyDSL-dominant kernels on more
operators than it fully passes, the signature of target-DSL dominance rather than
fallback. The gap is the measurable footprint of specification
shortcutting, whose mechanism we revisit in \secref{sec:models-errors}.

\subsection{Hardness Is a Property of the Operator, Not the Model}
\label{sec:models-hardness}

Averaging each operator's shape-level pass rate across the evaluated models
separates operator difficulty from model skill, yielding a ranking that no single
model's results would reveal (\tabref{tab:hardness}). The two ends are far
apart: nine operators are solved by every model, while the hardest,
\code{fp8\_blockscale\_fused\_moe}, clears only $22.2\%$ of attempts. The
hardest cluster is not random---\code{fp8\_blockscale\_fused\_moe} ($22.2\%$),
\code{fused\_rmsnorm\_quant} ($34.8\%$), \code{per\_token\_group\_quant\_fp8}
($44.7\%$), and \code{block\_scaled\_mm} ($56.9\%$) are all low-precision
quantization fused with a second operation, marking a shared capability frontier
rather than a weakness specific to any one model and pointing data curation at
fused fp8 kernels. \code{attention\_forward} ($45.2\%$) joins them from the
other direction, as the compute-bound case whose roof demands matrix-engine
scheduling (\secref{sec:models-regime}). No operator is failed by every
model, so the frontier is hard but not yet out of reach.

\begin{table}[t]
    \centering
    \begin{minipage}[t]{0.47\linewidth}
        \vspace{0pt}
        \centering
        \captionof{table}{Operator difficulty: mean shape-pass rate across the six
        models (hardest operators shown; the easiest nine pass on all).}
        \label{tab:hardness}
        {\footnotesize
        \setlength{\tabcolsep}{3pt}
        \renewcommand{\arraystretch}{1.15}
        \begin{tabular}{l S[table-format=3.1]}
            \toprule
            {Operator (hardest first)} & {Pass (\%)} \\
            \midrule
            \code{fp8\_blockscale\_fused\_moe}   & 22.2 \\
            \code{fused\_rmsnorm\_quant}         & 34.8 \\
            \code{per\_token\_group\_quant\_fp8} & 44.7 \\
            \code{attention\_forward}            & 45.2 \\
            \code{block\_scaled\_mm}             & 56.9 \\
            \emph{easiest 9 operators}             & 100.0 \\
            \bottomrule
        \end{tabular}}
    \end{minipage}\hfill%
    \begin{minipage}[t]{0.49\linewidth}
        \vspace{0pt}
        \centering
        \captionof{table}{Generation volume vs.\ quality
        (\secref{sec:models-cost}): output tokens, correct operators, and tokens
        per correct operator.}
        \label{tab:cost}
        {\footnotesize
        \setlength{\tabcolsep}{3pt}
        \renewcommand{\arraystretch}{1.15}
        \begin{tabular}{l S[table-format=1.2] S[table-format=2.0] S[table-format=3.0]}
            \toprule
            {Cand.} & {Out (M)} & {Corr.} & {Tok/op (K)} \\
            \midrule
            Claude Opus 4.7 & 5.67 & 24 & 236 \\
            GPT-5.5         & 1.19 & 26 & \bfseries 46 \\
            Qwen3.7-Max     & 2.02 & 23 & 88 \\
            Kimi-K2.6       & 2.75 & 23 & 120 \\
            GLM-5.1         & 1.84 & 12 & 153 \\
            DeepSeek-V4-Pro & 6.56 & 18 & 365 \\
            \bottomrule
        \end{tabular}}
    \end{minipage}
\end{table}

The full $30$-operator ranking, with production importance weights and per-operator
achievement, is in Appendix~\ref{app:per-op} (\tabref{tab:app-ops}).

\subsection{Where Utilization Goes: Memory- vs.\ Compute-Bound}
\label{sec:models-regime}

\begin{figure}[t]
    \centering
    \begin{minipage}[t]{0.48\linewidth}
        \vspace{0pt}
        \centering
        \includegraphics[width=\linewidth]{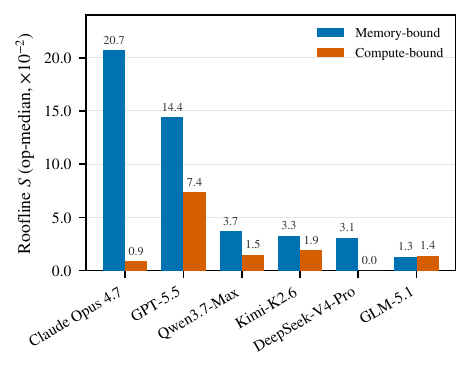}
        \caption{Per-model roofline achievement $S$ by regime (memory- vs.\
        compute-bound operators; $S$ from \eqnref{eq:per-op}).}
        \label{fig:regime}
    \end{minipage}\hfill%
    \begin{minipage}[t]{0.48\linewidth}
        \vspace{0pt}
        \centering
        \includegraphics[width=\linewidth]{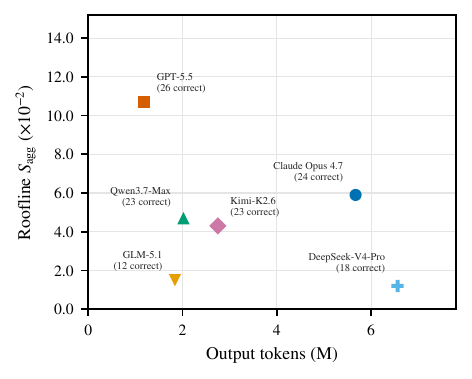}
        \caption{Output-token volume vs.\ production-weighted achievement
        $S_{\text{agg}}$ per model (labels give fully-correct operator counts).}
        \label{fig:cost}
    \end{minipage}
\end{figure}

Splitting operators by their semantic arithmetic intensity
($\text{AI}=W_{\text{flops}}/Q_{\text{bytes}}$, reference-derived, with a
$10$~FLOP/byte cut and the per-operator AI defined as the median across its shapes)
gives $16$ memory-bound and $10$ compute-bound operators (four
pure-indexing operators with no arithmetic are excluded), and the two regimes
behave differently (\figref{fig:regime}). Every backend reaches a far larger
fraction of the roof when the operator is bandwidth-bound, where a correct
memory-saturating kernel is comparatively easy to write, than when it is
compute-bound and demands tiling and software pipelining. The gap is starkest for
Opus~4.7, whose memory-bound median ($0.207$) is over twenty times its compute-bound
median ($0.009$); GPT-5.5 is the only model to reach a meaningful compute-bound
fraction ($0.074$), and that single difference is what lifts it above Opus~4.7 on
the production-weighted $S_{\text{agg}}$, since the heaviest operators are
compute-bound. The remaining backends sit far lower in both regimes.

The per-operator scores make the split concrete. On a bandwidth-bound reduction
such as \code{moe\_sum\_reduce}, Opus~4.7 reaches about three-fifths of the
memory roof ($S\!\approx\!0.59$); across the compute-bound operators its median
collapses to $0.009$, whereas GPT-5.5 holds a non-trivial fraction there
(median $0.074$, e.g.\ $0.22$ on the fused MoE GEMM). The same ordering holds for most backends, which
suggests the bottleneck is not correctness but scheduling: the agents can saturate
bandwidth but not the matrix engines. \figref{fig:regime} shows this
directly---for every model the memory-bound median towers over its compute-bound
median, and only GPT-5.5 lifts a compute-bound bar off the floor. The
reading is consistent across the panel:
agents have learned bandwidth optimization more thoroughly than compute
scheduling, and it argues for curating compute-bound, tile-and-pipeline kernels.

\subsection{Generation Volume Does Not Predict Quality}
\label{sec:models-cost}

Generation volume---the output tokens an agent emits to reach its final
candidate---does not predict quality (\tabref{tab:cost}). The two heaviest
generators land at opposite ends: DeepSeek-V4-Pro emits the most output ($6.56$M
tokens) yet is correct on only $18$ operators, whereas Opus~4.7 emits nearly as
much ($5.67$M) and turns it into $24$ correct operators and the most DSL-native
kernels. The two lightest split the same way: GPT-5.5 reaches the most correct
operators ($26$) on the fewest tokens ($1.19$M, $46$K per correct operator),
while GLM-5.1, on a comparable budget ($1.84$M), produces the fewest ($12$). More
generation implies neither a written kernel nor a correct one. \figref{fig:cost}
plots the two axes against each other---output-token volume against the
production-weighted achievement $S_{\text{agg}}$---and finds no relationship: the
lightest generator tops the panel while the heaviest sits near the bottom.

% fig:cost is paired with fig:regime, and tab:cost with tab:hardness, in
% \secref{sec:models-regime} (figures grouped with figures, tables with tables,
% so captions are consistent within each row).

\subsection{Error Modes and the Anti-Hacking Contract}
\label{sec:models-errors}

Classifying every failing unit across the evaluated models---$683$ in
total---shows both how generated kernels break and what the evaluation contract
keeps out.

\PHB{Compilation does fail, and it concentrates in the weaker backends.}
The per-shape gate isolates $365$ units ($53.4\%$ of all failures) that never
produce a compiled artifact (\tabref{tab:errors})---failures an import-level
gate would pass through to correctness. They track generation quality: GLM-5.1
and DeepSeek-V4-Pro compile only $60.9\%$ and $81.0\%$ of shapes, against $100\%$
for GPT-5.5 (\tabref{tab:main-xpu-a}), with GLM-5.1 alone accounting for $232$
of the $365$. The failures span the pipeline: syntax-invalid candidates,
exceptions on the first forward, ahead-of-time MLIR compilation that exceeds its
time budget, and structurally broken modules.

\PHM{Most surviving failures are silent.} Among kernels that do compile,
numeric mismatch is by far the most common failure mode ($257$ units, $37.6\%$ of all
failures): the kernel runs and returns a tensor of the
right shape and dtype but computes the wrong values. This is the regime only a
multi-seed numerical check can catch; a compile-or-run gate would pass it. The
remaining runtime failures are a tail of kernels that exceed the execution timeout
or raise an exception after a successful compile, alongside $24$ correct-but-slow
kernels that exceed the performance budget.

\begin{table}
    \caption{Failure taxonomy over all $683$ failing units across the six
    candidates, grouped into compile, runtime, and performance stages.}
    \label{tab:errors}
    \centering
    \small
    \renewcommand{\arraystretch}{1.15}
    \setlength{\tabcolsep}{8pt}
    \begin{tabular}{l l S[table-format=3.0] S[table-format=2.1]}
        \toprule
        {Failure category} & {What it is} & {Count} & {\%} \\
        \midrule
        \emph{Compile} & no compiled artifact produced & 365 & 53.4 \\
        \quad exception before first output & raised on import or first forward & 261 & 38.2 \\
        \quad compile timeout / OOM-kill & over the AOT compile budget & 85 & 12.4 \\
        \quad \code{Model} not \code{nn.Module} & not a valid \code{nn.Module} & 19 & 2.8 \\
        \midrule
        \emph{Runtime} & kernel ran, then failed & 294 & 43.0 \\
        \quad numeric mismatch & right shape and dtype, wrong values & 257 & 37.6 \\
        \quad runtime timeout & exceeded the execution budget & 26 & 3.8 \\
        \quad candidate-raised exception & error after a clean compile & 11 & 1.6 \\
        \midrule
        \emph{Performance} & over the time budget & 24 & 3.5 \\
        \quad correct-but-slow timeout & passed numerics, too slow & 24 & 3.5 \\
        \bottomrule
    \end{tabular}
\end{table}

\PHM{Hacking is prevented by construction, and what remains is made
visible.} The generation workspace never sees \code{metadata.json} (upstream
symbol, provenance, and production timing) or \code{roofline.json} (the SOL and
W/Q bounds), so a model can neither retrieve the reference implementation by name
nor fit its output to a scoring formula it is never shown. What this isolation
leaves available is specification shortcutting---satisfying the PyTorch reference
with a semantically equivalent but non-DSL implementation---and the FlyDSL-adoption
metric is what turns that from an invisible pass into a recorded number. The three
recurring patterns are PyTorch fallback (e.g.\
\code{scaled\_dot\_product\_attention} for attention), precompiled vendor-kernel
calls (e.g.\ AITER GEMM and MoE entry points), and DSL substitution (a Triton
kernel where FlyDSL was the target). The correctness illusion of
\secref{sec:models-illusion} is the aggregate of exactly these patterns: high
correctness paired with low FlyDSL adoption is the signature of a model optimizing
for the reference's input--output behavior rather than for writing the kernel.
The per-model breakdown of all failure modes is in Appendix~\ref{app:per-op}
(\tabref{tab:app-errors}).

\PHM{Takeaway.} Across candidates the empirical picture is consistent:
vanilla agents, even frontier models, leave substantial production-weighted roofline
performance on the table. Some failures are
hard---compilation that times out or crashes, numerically wrong output---but the
dominant soft failures are kernels that are correct yet reach less than an
eighth of the production-weighted roof, and kernels that pass without writing the
target DSL at all. The soft failures point less at raw coding ability than at
optimization knowledge that exists in expert practice but has not been distilled
into a form a base model can retrieve at generation time: tiling strategies
matched to a kernel's arithmetic intensity, vendor-specific memory-access
patterns, and dtype-aware fusion recipes. Closing that gap without per-task
retraining, by making that knowledge retrievable rather than relearned, is the
question the next section takes up.

\section{\atrexagent (\aka)}
\label{sec:skills}

\subsection{Motivation}
\label{sec:skills-motivation}

\secref{sec:models-errors} leaves two gaps that a stronger base model alone does
not close. The first is the \emph{correctness illusion}: a model can satisfy the
reference without writing the target DSL, passing while it spends almost no time in
FlyDSL (\secref{sec:models-illusion}). The second is the residual roofline gap:
kernels with FlyDSL-dominant execution still reach only a fraction of the
hardware roof. Both trace to missing GPU-optimization knowledge---roofline reasoning,
hardware-specific instruction choices, tiling, and accumulated tactics that frontier
models have seen only sparsely in training. \aka addresses both gaps by
combining a profile-driven optimization workflow with a curated GPU-optimization
knowledge base (GPU Wiki) that the workflow consults at each iteration. Two principles
run through the design: every hardware-spec value is sourced from GPU Wiki and archived
with its citation (no fabricated specs), and
every kernel change is justified by evidence from an official profiler rather than by intuition.

\subsection{Architecture and Workflow}
\label{sec:skills-taxonomy}

\begin{figure}[t]
    \centering
    \includegraphics[width=\linewidth]{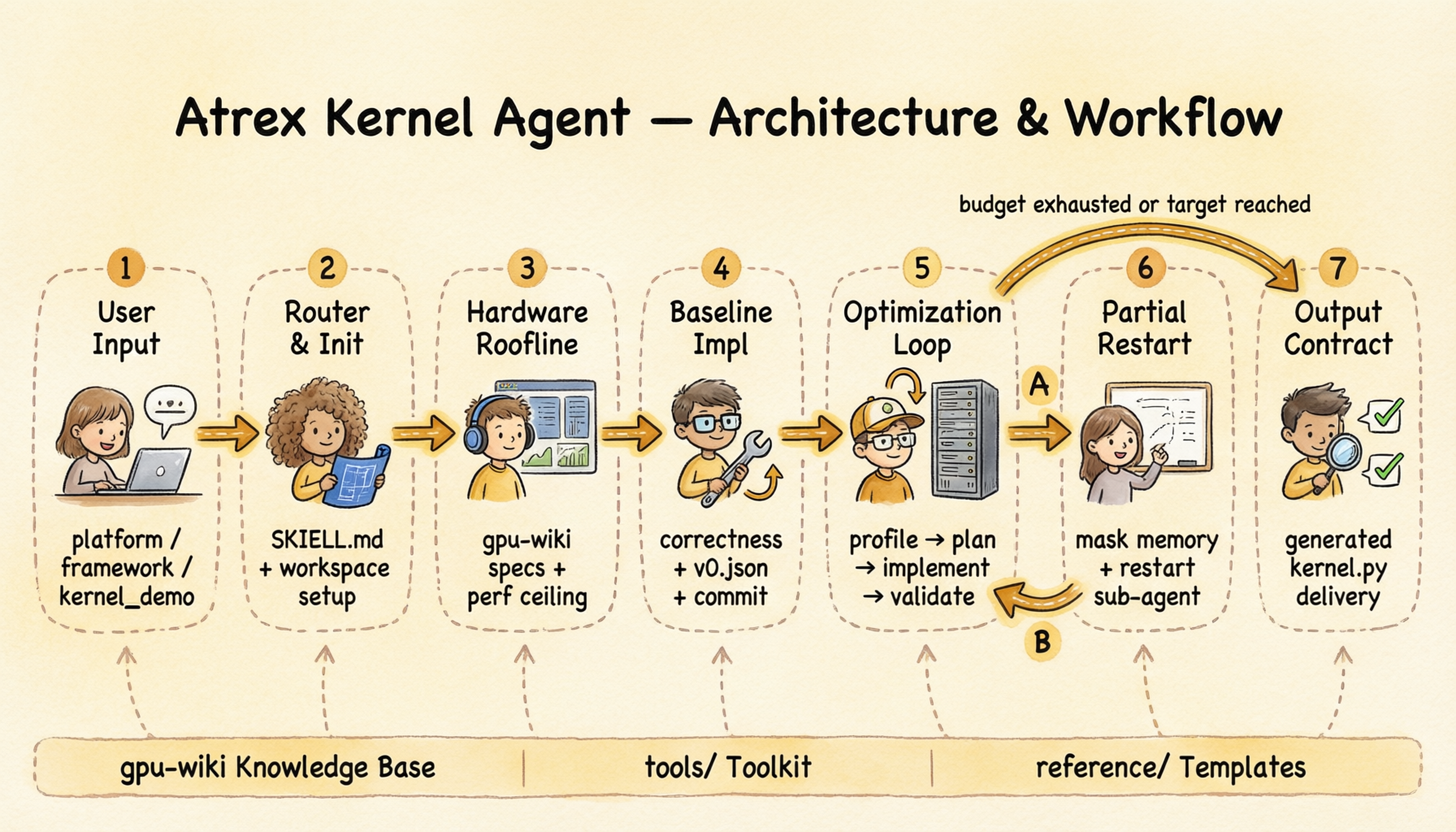}
    \caption{\aka architecture and outer workflow: the seven numbered stages from
    user input to evaluator-ready output, over shared resources (GPU Wiki,
    profiling and measurement tools, reference templates).}
    \label{fig:aka-architecture}
\end{figure}

\aka is not a single prompt template; it is a routed workflow whose intermediate
states are materialized as structured artifacts, profiles, and memory records
(\figref{fig:aka-architecture}). The design separates \emph{setup}, \emph{kernel
authoring}, \emph{profile-guided search}, \emph{partial restart}, and \emph{packaging} so that
the agent cannot skip hardware grounding, cannot silently overwrite failed attempts, and
always leaves an auditable path from the user's task to the submitted
kernel artifact. The seven numbered boxes in \figref{fig:aka-architecture}
define the outer workflow.

\PHB{\circledtext{1} User input.} The user supplies the
platform, framework or target DSL, and a kernel demo. Keeping this interface narrow is
intentional: the agent receives a concrete operator target rather than a broad instruction
to ``optimize code,'' which makes subsequent hardware lookup, harness construction, and
validation easier to scope to a single deployable kernel.

\PHM{\circledtext{2} Router and initialization.} The top-level
router parses the request, selects the relevant workflow components, creates an isolated
workspace, and records the task configuration, assumptions, and stop conditions in
the run state. This step gives the run a stable artifact boundary: later agents must
update the recorded state rather than relying on conversation-only memory.

\PHM{\circledtext{3} Hardware Roofline.} Before generating a
kernel, the router queries GPU Wiki for sourced hardware specifications, computes
FLOPs, bytes moved, arithmetic intensity, bound type, and the relevant Roofline ceiling,
and archives the source path for every hardware value. This grounding prevents two common
failure modes: optimizing for the wrong bottleneck and inventing peak throughput or
bandwidth numbers that make performance claims unverifiable.

\PHM{\circledtext{4} Baseline implementation.} The baseline
component builds the first correct target-DSL implementation and its test harness,
measures baseline latency and utilization, and initializes the run memory. The
baseline is deliberately correctness-first: it establishes a runnable kernel and a
rollback point before performance search starts.

\PHM{\circledtext{5} Optimization loop.} The profile optimizer
runs the inner loop expanded in \figref{fig:agent-loop}: profile, extract bottlenecks,
query knowledge, plan, apply one optimization category, validate, record memory, and
check whether to stop. Making optimization a loop rather than a one-shot rewrite lets the
agent attach each edit to profiler evidence and measure whether the edit actually helped.

\PHM{\circledtext{6} Partial restart.} When the optimizer concludes that no
actionable optimization direction remains, the partial-restart path masks a subset of iteration
memories while preserving the baseline, latest accepted kernel, and full audit trail,
then starts a fresh sub-agent from the remaining state. The partial view prevents prior
failed hypotheses from anchoring the new agent's judgment, allowing it to reassess the
kernel and identify opportunities that the previous search considered exhausted. This is
a \emph{dropout-style partial restart}: it drops selected search memories, not accepted
code or measured evidence, to help the optimizer escape a local optimum.

\PHM{\circledtext{7} Output contract.} The packaging
component converts the accepted implementation into a single evaluator-ready kernel
artifact. Tests, logs, notebooks, profiles, and external file dependencies are
excluded, so the final artifact is deployable code rather than a workspace snapshot.

\PHM{Execution components.} The system comprises five components under the
router: the baseline implementer, bottleneck-analysis helper, profile optimizer, partial
restart handler, and output-contract packager. Their separation is a control mechanism,
not merely an implementation convenience. Hardware reasoning stays in the bottleneck
helper, source-code edits stay in the implementer / optimizer, recovery is explicit
rather than implicit, and packaging is isolated from experimentation so that temporary
files cannot leak into the submitted artifact.

\PHM{GPU Wiki knowledge base.} The local knowledge base has two parts.
\emph{Reference kernels}: \textbf{298} reference kernels organized by vendor
architecture (\textbf{115} NVIDIA, \textbf{142} AMD, \textbf{41} architecture-agnostic),
which the agent can inspect or adapt. \emph{Documents}: \textbf{244} structured
optimization-knowledge documents, indexed by a relationship graph and organized into
five categories: optimization patterns (87; vendor-agnostic theory plus AMD- and
NVIDIA-specific details), framework/profiler/ISA references (109), DSL conversion
recipes (27), pitfall records (16), and hardware-spec tables (3). Hardware-spec
values used in a run must be sourced from GPU Wiki and archived with a path citation;
unsourced compute peaks, bandwidths, occupancy limits, or cache/LDS capacities
invalidate the run.

\PHM{Tools and external references.} The workflow also ships profiling and
measurement helpers for official-profiler data collection, kernel timing, utilization
analysis, and bandwidth-ceiling measurement. Beyond the local knowledge base, the
workflow can cache upstream reference projects (e.g., CUTLASS / CuTeDSL, FlyDSL,
Triton, AITER, FlashInfer, and FlashMLA) for source-level API and ISA lookup. Public
web lookup is a last resort when the local Wiki and reference projects are silent on
a low-level detail.

\subsection{Profile-Driven Optimization Loop}
\label{sec:skills-consumption}

\begin{figure}[t]
    \centering
    \includegraphics[width=\linewidth]{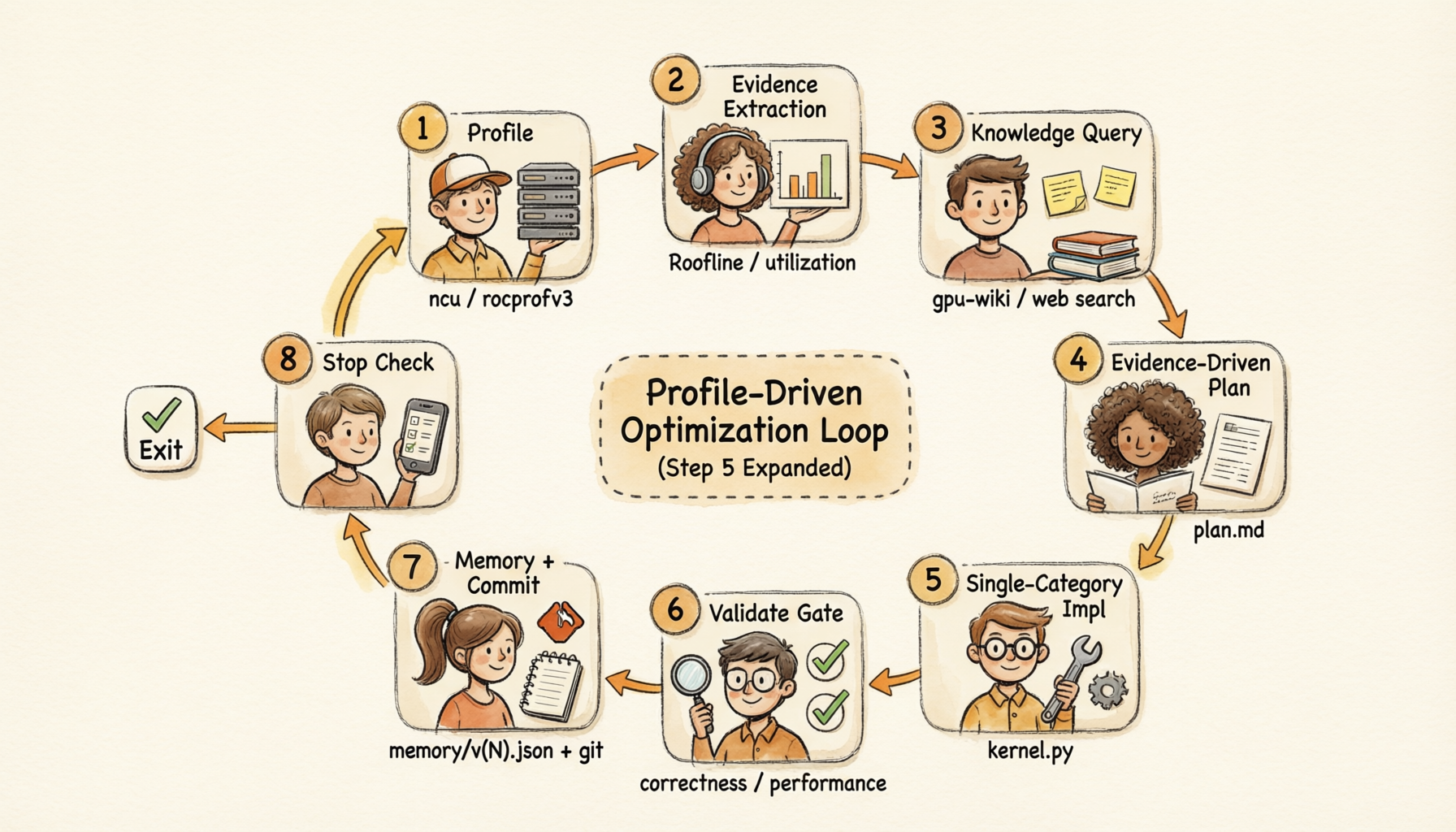}
    \caption{Profile-driven optimization loop inside \aka: the eight numbered stages
    that turn profiler evidence into one scoped, validated implementation change per
    iteration.}
    \label{fig:agent-loop}
\end{figure}

The outer workflow enters the optimization loop only after a hardware Roofline estimate and a
correct baseline exist. \figref{fig:agent-loop} expands Step~5 of
\figref{fig:aka-architecture} into eight ordered stages. The order matters: each iteration
starts from measurement, turns measurement into a hypothesis, maps the hypothesis to a
documented optimization tactic, changes one class of code, and records the outcome before
deciding whether another iteration is justified.

The eight stages are:

\PHB{\circledtext{1} Profile.} The loop profiles the current
kernel with the official hardware profiler---\code{ncu} on NVIDIA or
\code{rocprofv3}/ATT/PMC/assembly on AMD---rather than using latency alone. Timers still
determine whether a change is faster, but profiler counters determine what kind of
bottleneck the next edit should target.

\PHM{\circledtext{2} Evidence extraction.} The bottleneck helper
converts raw counters into Roofline and utilization evidence: achieved bandwidth or
compute, occupancy, memory transactions, wave / warp behavior, instruction mix, and
whether the observed limit matches the semantic bound computed in the outer workflow. The
output is a concrete bottleneck hypothesis, not a generic request for speedup.

\PHM{\circledtext{3} Knowledge query.} The agent retrieves the
relevant GPU Wiki notes, hardware-spec entries, pitfalls, reference kernels, and
local reference-project source. Public web lookup is allowed only when the local
knowledge base and cached projects are silent. This step keeps optimization tactics
grounded in documented hardware and library behavior.

\PHM{\circledtext{4} Evidence-driven plan.} The agent writes an explicit plan:
the measured bottleneck, the knowledge sources it used, the expected
performance effect, the correctness risks, and the single optimization category to try
next. Requiring an explicit plan forces the model to connect evidence to action before
editing code.

\PHM{\circledtext{5} Single-category implementation.} The agent
edits the candidate kernel in one category at a time: tiling, vectorization, memory layout,
prefetching, synchronization, split / fusion structure, or instruction selection. This
constraint trades off search speed for attribution: when performance changes, the log can
identify which class of edit caused it.

\PHM{\circledtext{6} Validate gate.} Correctness is checked
before performance. A kernel that fails correctness is rejected immediately; a correct
kernel is timed under the same harness and compared with the previous accepted state.
Regressions are reverted unless they are explicitly kept as a measured trade-off toward a
later plan.

\PHM{\circledtext{7} Memory update.} Accepted iterations update the structured
run memory and running summary; rejected attempts are recorded with the evidence and
failure reason. This separates search from memory: later iterations reason over
structured records instead of relying on hidden conversation state.

\PHM{\circledtext{8} Stop check.} The loop does not stop by a
fixed external rule such as a hard target or exhausted budget. Instead, the agent inspects
the current measurement history, accepted and rejected edits, and unmasked memory, then
decides whether it has achieved the objective well enough or whether no actionable
optimization direction remains. In the latter case, control returns to the outer
partial-restart path (\figref{fig:aka-architecture}), which masks selected memories
before a fresh sub-agent reassesses the remaining optimization space.

This artifact boundary is what makes the loop more than repeated prompting. Each
iteration must turn profiler evidence into a concrete bottleneck hypothesis, retrieve the
relevant GPU Wiki notes or reference-project source, write an explicit plan, and
change exactly one optimization category before validation. The resulting log gives the
case studies below a causal trace: not just that the final kernel is faster, but which
evidence led to which edit and which edits were kept or reverted.

In the evaluation below (\secref{sec:skills-eval-setup}), the \emph{vanilla} condition
uses the standard one-pass FlyDSL prompt (\secref{sec:models-setup}); the \emph{\aka}
condition runs the same base model inside this workflow. Everything else is held fixed:
base model, target DSL, evaluation contract, and the hidden-information policy of
\secref{sec:models}.

\subsection{Optimizer-Augmented Evaluation: Setup}
\label{sec:skills-eval-setup}

\PHB{Scope.} This is a controlled case study, not a benchmark-wide ranking. We
pair two base models from the \secref{sec:models} panel---Claude Opus~4.7 and
Qwen3.7-Max---with the two conditions, on three operators that stress the agent in
different ways. By semantic arithmetic intensity all three are compute-bound
(\secref{sec:models-regime}), but the bottleneck they actually hit differs:
\code{attention\_forward} is throughput-bound and can approach the matrix-engine roof,
whereas \code{chunk\_gated\_delta\_rule\_state} (a linear-attention recurrence) and
\code{mla\_decode\_attention} run at small, dispatch-bound shapes where roofline
achievement stays near the floor and the deployed kernel's latency is the meaningful
target. Each operator uses one production-representative shape, and both conditions run on
that same shape and the same XPU-A hardware. Metrics follow \secref{sec:models-setup}.

\PHM{One extra signal.} The optimizer-augmented condition is a loop, so it leaves an
iteration log: each profiled hypothesis and its measured result. The case studies read
this log to explain \emph{why} a kernel improved---evidence the aggregate numbers cannot
give.

\PHM{The headline metric depends on the regime.} No single number reads the same
across operators. For the throughput-bound \code{attention\_forward} we report $S$, the
fraction of the roof. For the recurrence, where the vanilla failure is a PyTorch fallback,
we report the jump in FlyDSL adoption together with the speedup. For the dispatch-bound
\code{mla\_decode\_attention}, $S$ saturates near the floor, so we report the ratio to
the production kernel.

\PHM{Data status.} All three operators are complete in both conditions for both
base models. Performance is a single timed run per (op, shape).
\tabref{tab:skills-eval} gives the comparison.

\subsection{Main Results}
\label{sec:skills-main}

The two base models expose two distinct effects of the workflow. On the weaker
Qwen3.7-Max the effect is \emph{fallback-to-target-DSL dominance}. Its vanilla kernels barely
use FlyDSL---$0\%$ on the recurrence and \code{attention\_forward}, $14\%$ on
\code{mla\_decode}, passing by PyTorch fallback (\secref{sec:models-illusion})---while
the optimizer-augmented agent writes near-100\% FlyDSL kernels that overtake production
(\tabref{tab:skills-eval}). The recurrence \code{chunk\_gated\_delta\_rule\_state} is
the clearest case: a $0\%$-FlyDSL fallback roughly $29\times$ slower than production
becomes a near-100\%-FlyDSL kernel that beats it ($1.2\times$). On the throughput-bound
\code{attention\_forward} it gains the most in roofline terms ($S$ $0.06 \to 0.40$,
$6.7\times$ over the fallback) and edges past the hand-tuned AITER kernel
($1.11\times$); on the dispatch-bound \code{mla\_decode} it likewise passes production
($1.06\times$). All three optimizer-augmented kernels overtake the deployed kernel, whereas all three
vanilla kernels trailed it ($0.03\times$ to $0.17\times$).

On the stronger Opus~4.7 the effect is mostly \emph{optimization}: on the two attention
operators its vanilla kernels are already FlyDSL (${\sim}100\%$ on
\code{attention\_forward}, $92\%$ on \code{mla\_decode}), so the workflow pushes
already-written kernels toward the roof---lifting \code{attention\_forward} from
$S=0.28$ to $0.42$ ($0.78\times \to 1.17\times$ production) and carrying
\code{mla\_decode} from $0.71\times$ to $1.27\times$. The recurrence is the exception:
like Qwen, Opus falls back to PyTorch in the vanilla condition ($0\%$ FlyDSL), and the
workflow converts it into a near-100\%-FlyDSL kernel that beats production
($0.04\times \to 1.31\times$). Read together, the two models trace the same library to two
complementary roles: for a weaker base model it closes the \emph{target-DSL-dominance} gap broadly
(fallback~$\to$~FlyDSL), and for a stronger one it closes the residual \emph{roofline} gap
on already-written kernels while still converting the one recurrence where even it falls
back---beating the hand-tuned production kernel on every operator.

\begin{table}
    \caption{Optimizer-augmented evaluation on Qwen3.7-Max and Claude Opus~4.7 (one
    production shape per operator, XPU-A). Each cell reads
    vanilla~$\to$~optimizer-augmented for FlyDSL share, roofline $S$, and speedup
    over the production kernel.}
    \label{tab:skills-eval}
    \centering
    \small
    \renewcommand{\arraystretch}{1.25}
    \setlength{\tabcolsep}{6pt}
    \begin{tabular}{l l c c c}
        \toprule
        & & \multicolumn{3}{c}{vanilla $\to$ optimizer-augmented} \\
        \cmidrule(lr){3-5}
        Model & Operator & FlyDSL & $S$ (roofline) & vs.\ prod \\
        \midrule
        Qwen3.7-Max & \code{chunk\_gated\_delta} & $0\% \to {\sim}100\%$ & $0.001 \to 0.03$ & $0.03\times \to 1.20\times$ \\
                    & \code{attention\_forward} & $0\% \to 99\%$ & $0.06 \to 0.40$ & $0.17\times \to 1.11\times$ \\
                    & \code{mla\_decode} & $14\% \to 87\%$ & $0.0003 \to 0.0035$ & $0.10\times \to 1.06\times$ \\
        \midrule
        Claude Opus 4.7 & \code{chunk\_gated\_delta} & $0\% \to {\sim}100\%$ & $0.001 \to 0.03$ & $0.04\times \to 1.31\times$ \\
                 & \code{attention\_forward} & ${\sim}100\% \to 99\%$ & $0.28 \to 0.42$ & $0.78\times \to 1.17\times$ \\
                 & \code{mla\_decode} & $92\% \to 92\%$ & $0.0023 \to 0.0042$ & $0.71\times \to 1.27\times$ \\
        \bottomrule
    \end{tabular}
\end{table}

\subsection{Case Studies: How the Optimization Workflow Helps}
\label{sec:skills-cases}

The aggregate results show the workflow's effect; the iteration logs show how each
improvement arises. We examine one operator per regime, and each case ends with the
specific advantage it demonstrates.

\subsubsection{\code{chunk\_gated\_delta\_rule\_state}: from fallback to beating production}
\label{sec:skills-case-chunk}

This recurrence is the clearest case. In the vanilla condition, Qwen3.7-Max returns a
correct module that is pure PyTorch fallback---0\% FlyDSL---and, at about $2.8$~ms, is
roughly $29\times$ slower than the production kernel ($\approx\!95~\mu$s). A base model
has no learned FlyDSL template for this recurrence, so it composes framework ops instead.

The optimizer-augmented workflow retrieves a direct scaffold: the AITER FlyDSL
chunk-gated-delta reference kernel, a Wiki playbook for this operator family on XPU-A,
the hardware-spec sheet, and a pitfalls entry (raw vs.\ cumulative gates, the state
layout, and the 80-CU occupancy limit). The agent adapts the scaffold into FlyDSL, then
optimizes against profiles: a tile sweep up to the LDS limit, prefetch and store
reordering, and padding to remove bank conflicts, reverting each regression with a
logged reason. The result is a near-100\% FlyDSL kernel that runs in about $79~\mu$s,
approximately $1.2\times$ faster than the production kernel. Because this small-shape recurrence is
dispatch-bound, roofline achievement stays near the floor ($S\!\approx\!0.03$), so the
signal here is the target-DSL-dominance jump and the speedup rather than the fraction of the roof.
\code{torch.compile} does not rescue the recurrence either---it stays a PyTorch
composition at ${\approx}3$~ms, essentially the vanilla fallback---so the FlyDSL kernel is
about $38\times$ faster than it as well.

The advantage is concrete. The reference kernel gave the model a way to write the
recurrence in FlyDSL, and the profile loop turned a first correct version into a faster
one. This is the correctness illusion of \secref{sec:models-illusion}, closed in one
step. Opus~4.7 shows the same conversion on this recurrence: its vanilla kernel is
likewise a $0\%$-FlyDSL fallback, and the optimizer-augmented kernel reaches near-100\%
FlyDSL at $1.31\times$ production---even the stronger base model falls back here and is
rescued by the same scaffold.

\subsubsection{\code{attention\_forward}: the same conversion, compute-bound}
\label{sec:skills-case-attn}

The compute-bound operator shows the same conversion, with a smaller margin over
production. Vanilla Qwen3.7-Max again writes a correct but 0\%-FlyDSL kernel; the
optimizer-augmented agent writes a 99\%-FlyDSL kernel that runs $6.7\times$ faster, raising
$S$ from $0.06$ to $0.40$ and edging past the hand-tuned AITER production
kernel ($1.11\times$). That a generated kernel overtakes AITER here is less
surprising than it first reads: on this shape---a single $65{,}556$-token sequence with
$16$ heads and a head dimension of $72$, an awkward non-power-of-two for MFMA
tiling---the general AITER varlen kernel itself reaches only ${\approx}36\%$ of
the roof ($S \approx 0.36$), leaving room for a shape-specialized kernel to pass it. The
library's value is on the compute path: the workflow picks among its FlyDSL
flash-attention references, avoids documented construction pitfalls that otherwise cause
repeated compile failures, and follows the ISA targets---vectorized access, MFMA
utilization, no register spills. The stronger base model sharpens the same point:
Opus~4.7's vanilla kernel is already FlyDSL at $S=0.28$, and the workflow lifts it to
$S=0.42$ at $1.17\times$ production---optimization rather than conversion, but the same
lever set. The advantage here is direction: the library tells the agent which levers
move a working kernel toward the roof.

\subsubsection{\code{mla\_decode\_attention}: edging past hand-tuned production}
\label{sec:skills-case-mla}

The latency-bound operator is the hardest test, because its ceiling is a hand-written
assembly kernel rather than a framework op. The vanilla agent stays far from that ceiling:
a mostly-fallback kernel at about $0.1\times$ the production AITER time. The
optimizer-augmented workflow classifies the problem as dispatch-bound and restructures the
decode as a split-KV computation with an online-softmax reduction, spreading the scattered
gather across compute units. After profiling-driven tuning, the resulting $87\%$-FlyDSL
kernel reaches about $18~\mu$s and edges past the hand-tuned AITER kernel
($1.06\times$). This is the result that most
directly answers the paper's question: a generated kernel beating a hand-written
production kernel on a deployed operator. The advantage here is structural reasoning---the
split-KV restructuring the base model adopts only once the workflow frames the problem by
its bound.

\medskip
\noindent
Across all three operators, the advantage is similar in kind. \aka does not hand
the agent a finished kernel. It supplies the missing pieces---a regime-appropriate target,
a reference scaffold, the pitfalls to avoid, the ISA-level levers, and a measurement
loop---and the agent does the rest. These are the gaps \secref{sec:models} traced the
vanilla failures to.

\section{Discussion}
\label{sec:discussion}

\subsection{Limitations}
\label{sec:discussion-limitations}

\PHB{Trace scope and evaluation coverage.} Trace collection in this release covers selected XPU-A and H20 production clusters with more than 10k deployed cards. These traces cover compute-limited, memory-rich GPU fleets, so the sampled problems represent that traced deployment slice rather than every hardware class. This draft reports empirical results only on XPU-A; broader accelerator validation remains future work and will test whether the same benchmark construction and roofline-normalized scoring remain informative across accelerator families. We do not claim that the current trace slice represents every future hardware class.

\PHM{Inference-only.} The current release covers inference kernels exclusively. Training workloads---in particular backward kernels and optimizer-step kernels---are out of scope for this version.

\subsection{Future Work}
\label{sec:discussion-future}

\PHB{More accelerators and workload regimes.} Future releases should extend empirical
evaluation beyond the current XPU-A setting to NVIDIA-side deployments and
higher-throughput accelerators. This expansion is not only a hardware exercise:
different accelerator classes host different model mixes, precision formats,
kernel libraries, and serving or training paths, so the benchmark should grow by
capturing the workload regimes that appear on those fleets. The
roofline-normalized score generalizes to these settings; the main engineering
work is calibrating hardware peaks and validating the reference and evaluation
path per accelerator.

\PHM{Periodic benchmark refresh.} Production serving workloads change as models, engines, and kernel libraries evolve. We therefore plan to refresh \atrexbench on a regular cadence from new production traces, updating operator sets, shape distributions, and importance weights while keeping each released version frozen for reproducibility.

\PHM{Scaling agentic optimization.} Future versions of \aka will decompose kernels into prologue, mainloop, and epilogue regions for context-isolated sub-agent tuning, and explore multi-model planning, implementation, and review workflows. Longer campaigns of 300 or more iterations will distill validated experience back into GPU Wiki, while new templates, cross-project practices, and composable optimization primitives expand the search space. More precise retrieval and progressive disclosure will expose this knowledge incrementally as required by the search process.

\section{Conclusion}
\label{sec:conclusion}

We presented \textbf{\atrexbench}, a GPU kernel generation benchmark sourced from full-cluster production inference traces and scored with importance-weighted roofline metrics. On 30 operators and 440 shapes, six frontier coding agents reach at most ${\sim}10\%$ of the hardware roofline---with the residual gap concentrated in domain-knowledge-intensive patterns rather than raw coding ability. Motivated by this finding, we co-released \textbf{\aka}---a profile-driven kernel-optimization agent that combines iterative measure--revise search, optimization dropout for escaping stalled search contexts, and a layered knowledge base of expert experience, reference kernels, and upstream open-source practices. A controlled case study shows that the agent converts PyTorch fallbacks into real FlyDSL kernels that match or exceed hand-tuned production baselines. Both artifacts are open-source; we hope they provide a deployment-aligned evaluation signal and a reusable optimization agent that lowers the barrier to production-relevant progress in LLM kernel generation.

\bibliographystyle{plainnat}
\bibliography{references}

@article{williams2009roofline,
  title = {Roofline: An insightful visual performance model for multicore architectures},
  author = {Williams, Samuel and Waterman, Andrew and Patterson, David},
  journal = {Communications of the ACM},
  volume = {52},
  number = {4},
  pages = {65--76},
  year = {2009}
}

@article{ouyang2025kernelbench,
  title = {{KernelBench}: Can {LLMs} Write Efficient {GPU} Kernels?},
  author = {Ouyang, Anne and Guo, Simon and Arora, Simran and Zhang, Alex L.\ and Hu, William and R{\'e}, Christopher and Mirhoseini, Azalia},
  journal = {arXiv preprint arXiv:2502.10517},
  year = {2025}
}

@misc{saroufim2025backendbench,
  title = {{BackendBench}: An Evaluation Suite for Testing How Well {LLMs} and Humans Can Write {PyTorch} Backends},
  author = {Saroufim, Mark and Wang, Jiannan and Maher, Bert and Paliskara, Sahan and Wang, Laura and Sefati, Shahin and Candales, Manuel},
  howpublished = {\url{https://github.com/meta-pytorch/BackendBench}},
  year = {2025}
}

@inproceedings{li2025tritonbench,
  title = {{TritonBench}: Benchmarking Large Language Model Capabilities for Generating {Triton} Operators},
  author = {Li, Jianling and Li, Shangzhan and Gao, Zhenye and Shi, Qi and Li, Yuxuan and Wang, Zefan and Huang, Jiacheng and Wang, Haojie and Wang, Jianrong and Han, Xu and Liu, Zhiyuan and Sun, Maosong},
  booktitle = {Proc. ACL Findings},
  pages = {23053--23066},
  year = {2025}
}

@article{wen2025multikernelbench,
  title = {{MultiKernelBench}: A Multi-Platform Benchmark for Kernel Generation},
  author = {Wen, Zhongzhen and Zhang, Yinghui and Li, Zhong and Xie, Linna and Zhang, Tian and Liu, Zhongxin},
  journal = {arXiv preprint arXiv:2507.17773},
  year = {2025}
}

@article{xing2026flashinferbench,
  title = {{FlashInfer-Bench}: Building the Virtuous Cycle for {AI}-driven {LLM} Systems},
  author = {Xing, Shanli and Zhai, Yiyan and Jiang, Alexander and Dong, Yixin and Wu, Yong and Ye, Zihao and Ruan, Charlie and Huang, Yingyi and Zhang, Yineng and Yin, Liangsheng and Bayyapu, Aksara and Ceze, Luis and Chen, Tianqi},
  journal = {arXiv preprint arXiv:2601.00227},
  year = {2026}
}

@article{zhu2026cudabench,
  title = {{CUDABench}: Benchmarking {LLMs} for Text-to-{CUDA} Generation},
  author = {Zhu, Jiace and Chen, Wentao and Fan, Qi and Ren, Zhixing and Wu, Junying and Chai, Xing Zhe and Rungrueangwutthinon, Chotiwit and Ma, Yehan and Zou, An},
  journal = {arXiv preprint arXiv:2603.02236},
  year = {2026}
}

@article{lin2026solexecbench,
  title = {{SOL-ExecBench}: Speed-of-Light Benchmarking for Real-World {GPU} Kernels Against Hardware Limits},
  author = {Lin, Edward and Modi, Sahil and Hari, Siva Kumar Sastry and Huang, Qijing and Ye, Zhifan and Qin, Nestor and Zhou, Fengzhe and Zhang, Yuan and Wang, Jingquan and Damani, Sana and Peri, Dheeraj and Xie, Ouye and Kane, Aditya and Maor, Moshe and Behar, Michael and Cao, Triston and Mehta, Rishabh and Singh, Vartika and Mailthody, Vikram Sharma and Chen, Terry and Ye, Zihao and Chen, Hanfeng and Chen, Tianqi and Grover, Vinod and Chen, Wei and Liu, Wei and Chung, Eric and Ceze, Luis and Bringmann, Roger and Zeller, Cyril and Lightstone, Michael and Kozyrakis, Christos and Shi, Humphrey},
  journal = {arXiv preprint arXiv:2603.19173},
  year = {2026}
}

@inproceedings{kwon2023vllm,
  title = {Efficient Memory Management for Large Language Model Serving with {PagedAttention}},
  author = {Kwon, Woosuk and Li, Zhuohan and Zhuang, Siyuan and Sheng, Ying and Zheng, Lianmin and Yu, Cody Hao and Gonzalez, Joseph E. and Zhang, Hao and Stoica, Ion},
  booktitle = {Proc. ACM SOSP},
  pages = {611--626},
  year = {2023},
  doi = {10.1145/3600006.3613165}
}

@inproceedings{zheng2024sglang,
  title = {{SGLang}: Efficient Execution of Structured Language Model Programs},
  author = {Zheng, Lianmin and Yin, Liangsheng and Xie, Zhiqiang and Sun, Chuyue and Huang, Jeff and Yu, Cody Hao and Cao, Shiyi and Kozyrakis, Christos and Stoica, Ion and Gonzalez, Joseph E. and Barrett, Clark and Sheng, Ying},
  booktitle = {Proc. NeurIPS},
  year = {2024},
  note = {arXiv:2312.07104}
}

@misc{amd2025aiter,
  title = {{AITER}: {AI} Tensor Engine for {ROCm}},
  author = {{AMD}},
  howpublished = {\url{https://github.com/ROCm/aiter}},
  year = {2025},
  note = {High-performance AI operator library for ROCm.}
}

@misc{flydsl2025github,
  title = {{FlyDSL}},
  author = {{AMD}},
  howpublished = {\url{https://github.com/ROCm/FlyDSL}},
  year = {2025}
}

@article{tan2026rtpllm,
  title = {{RTP-LLM}: High-Performance {Alibaba} {LLM} Inference Engine},
  author = {Tan, Boyu and Guo, Jiarui and Lv, Zongwei and Sun, Hanbo and Yang, Tong and Liu, Kan and Shi, Xinfei and Hu, Zetao and Yu, Yaxin and Zhang, Chi and Zhang, Jianning and Yang, Xi and Zhang, Wei and Cai, Bo and Zhou, Silu and Wang, Xiyu and He, Na and Yu, Yinghao and Bao, Wending and Huang, Guiyang and Yuan, Yuxing and Yin, Juncheng and Wang, Nan and Yang, Lin and Zhang, Zechao and Chen, Lu and Li, Guoding and Lan, Tao and Qu, Lin},
  journal = {arXiv preprint arXiv:2605.29639},
  year = {2026}
}

@article{wang2023voyager,
  title = {{Voyager}: An Open-Ended Embodied Agent with Large Language Models},
  author = {Wang, Guanzhi and Xie, Yuqi and Jiang, Yunfan and Mandlekar, Ajay and Xiao, Chaowei and Zhu, Yuke and Fan, Linxi and Anandkumar, Anima},
  journal = {arXiv preprint arXiv:2305.16291},
  year = {2023}
}

@inproceedings{zhou2024selfdiscover,
  title = {Self-Discover: Large Language Models Self-Compose Reasoning Structures},
  author = {Zhou, Pei and Pujara, Jay and Ren, Xiang and Chen, Xinyun and Cheng, Heng-Tze and Le, Quoc V.\ and Chi, Ed H.\ and Zhou, Denny and Mishra, Swaroop and Zheng, Huaixiu Steven},
  booktitle = {Proc. NeurIPS},
  year = {2024},
  note = {arXiv:2402.03620}
}

@misc{anthropic2025skills,
  title = {Skills: Specialized capabilities for {Claude}},
  author = {{Anthropic}},
  howpublished = {\url{https://www.anthropic.com/news/skills}},
  year = {2025},
  note = {Anthropic product announcement; accessed for description of agent-side skill packaging.}
}

@misc{ako2026,
  title = {{AKO}: Agentic Kernel Optimization},
  author = {Xie, Shuxiao and Xie, Shuyang and Ran, Dezhi and Yang, Wei and Xie, Tao},
  howpublished = {\url{https://tongminglaic.github.io/AKO}},
  year = {2026},
  note = {Technical report.}
}

@misc{kda2026github,
  title = {{Kernel Design Agents}},
  author = {{MIT HAN Lab}},
  howpublished = {\url{https://github.com/mit-han-lab/kernel-design-agents}},
  year = {2026}
}

@inproceedings{yang2024sweagent,
  title = {{SWE-agent}: Agent--Computer Interfaces Enable Automated Software Engineering},
  author = {Yang, John and Jimenez, Carlos E.\ and Wettig, Alexander and Lieret, Kilian and Yao, Shunyu and Narasimhan, Karthik and Press, Ofir},
  booktitle = {Proc. NeurIPS},
  year = {2024},
  note = {arXiv:2405.15793}
}

\appendix

\section{Per-Operator Results}
\label{app:per-op}

This appendix expands the operator-balanced aggregates of \secref{sec:models}
into per-operator and per-model detail. All numbers come from the same evaluation
run as the main text---six candidates, $30$ operators, $440$ shapes, XPU-A---with
the metrics defined in \secref{sec:models-setup}.

\PHB{The production-weighted score is set by a few heavy, low-achievement
operators.} \tabref{tab:app-ops} lists every operator with its roofline regime,
median semantic arithmetic intensity across its shapes, production importance weight $w_i$, shape count, mean
shape-pass rate across the six models (the model-independent difficulty of
\secref{sec:models-hardness}), and the per-operator achievement $S_i$ of the two
strongest backends. Two facts drive $S_{\text{agg}}$. First, importance is
concentrated: the five heaviest operators hold $63.7\%$ of weighted production
wall-time, and \code{unified\_attention} alone holds $36.1\%$. Second, those heavy
operators are exactly the ones no model reaches---\code{unified\_attention} tops out
at $S=0.07$ (GPT-5.5) and $0.01$ (Opus~4.7), \code{fused\_moe} at $0.22$,
\code{block\_scaled\_mm} at $0.21$---while several light bandwidth-bound operators
clear $S>0.3$ (\code{moe\_sum\_reduce} $0.59$, \code{l2\_norm} $0.44$). Weighting by
$w_i$ therefore pulls the aggregate down to ${\sim}0.11$ and localizes the Opus--GPT
split of \secref{sec:models-main}: the two are even on the easy operators, but
GPT-5.5 is the only model with non-trivial $S$ on the heavy compute-bound ones
(\code{unified\_attention} $0.07$ vs.\ $0.01$; \code{fused\_moe} $0.22$ vs.\
unmeasured, as Opus passes it only by a non-DSL fallback). The difficulty column reads
the suite from the other direction: the hardest operators are uniformly low-precision
fused or matrix-engine kernels (\code{fp8\_blockscale\_fused\_moe} $22.2\%$,
\code{fused\_rmsnorm\_quant} $34.8\%$, \code{per\_token\_group\_quant\_fp8}
$44.7\%$, \code{attention\_forward} $45.2\%$), while nine elementwise and
normalization operators are solved by every model.

\begin{table}[p]
    \caption{All $30$ operators, sorted by importance weight $w_i$. AI is the
    median per-shape arithmetic intensity in FLOP/byte; Regime is its class
    (comp/mem/idx; ``--'' for
    pure-indexing); Pass is the mean shape-pass across the six models;
    $S_{\text{Opus}}$, $S_{\text{GPT}}$ are the per-operator median $S$ of the two
    leaders.}
    \label{tab:app-ops}
    \centering
    \scriptsize
    \setlength{\tabcolsep}{3pt}
    \renewcommand{\arraystretch}{1.15}
    \begin{tabular}{l c S[table-format=4.1] S[table-format=2.1] S[table-format=2.0] S[table-format=3.1] S[table-format=1.2] S[table-format=1.2]}
        \toprule
        {Operator} & {Regime} & {AI} & {$w_i$\,(\%)} & {\#sh} & {Pass\,(\%)} & {$S_{\text{Opus}}$} & {$S_{\text{GPT}}$} \\
        \midrule
        \code{unified\_attention} & comp & 112 & 36.1 & 25 & 69.3 & 0.01 & 0.07 \\
        \code{fused\_moe} & comp & 415 & 10.4 & 23 & 100.0 & {--} &0.22 \\
        \code{block\_scaled\_mm} & comp & 823 & 8.5 & 24 & 56.9 & 0.16 & 0.21 \\
        \code{fp8\_blockscale\_fused\_moe} & comp & 121 & 4.7 & 6 & 22.2 & {--} &0.02 \\
        \code{paged\_attention\_decode} & mem & 2.6 & 4.0 & 8 & 66.7 & 0.02 & 0.02 \\
        \code{reshape\_and\_cache} & idx & {--} &4.0 & 8 & 100.0 & 0.23 & 0.32 \\
        \code{topk\_filter} & idx & {--} &3.2 & 8 & 83.3 & 0.03 & 0.01 \\
        \code{gated\_delta\_rule\_update} & comp & 90 & 3.2 & 17 & 57.8 & 0.01 & 0.01 \\
        \code{fused\_qkv\_rope} & mem & 0.7 & 3.1 & 6 & 100.0 & 0.00 & 0.00 \\
        \code{rms\_norm} & mem & 0.8 & 2.5 & 56 & 83.3 & 0.17 & 0.07 \\
        \code{mla\_decode\_attention} & comp & 34 & 2.1 & 3 & 83.3 & 0.00 & 0.01 \\
        \code{attention\_forward} & comp & 4068 & 2.0 & 21 & 45.2 & 0.16 & 0.14 \\
        \code{causal\_conv1d} & mem & 3.0 & 2.0 & 9 & 81.5 & 0.07 & 0.07 \\
        \code{moe\_topk\_gating\_softmax} & mem & 0.9 & 1.9 & 10 & 91.7 & 0.01 & 0.01 \\
        \code{fused\_qk\_rmsnorm} & mem & 0.6 & 1.7 & 4 & 100.0 & 0.00 & 0.00 \\
        \code{silu\_and\_mul} & mem & 0.8 & 1.6 & 37 & 100.0 & 0.27 & 0.25 \\
        \code{chunk\_gated\_delta\_rule\_state} & comp & 45 & 1.4 & 25 & 67.3 & 0.00 & 0.00 \\
        \code{fused\_add\_rms\_norm} & mem & 0.5 & 1.0 & 19 & 83.3 & 0.30 & 0.21 \\
        \code{moe\_align\_block\_size} & idx & {--} &1.0 & 6 & 100.0 & 0.00 & 0.00 \\
        \code{mrope} & mem & 0.6 & 0.9 & 5 & 83.3 & 0.27 & 0.23 \\
        \code{chunk\_delta\_rule\_output} & comp & 51 & 0.8 & 16 & 50.0 & 0.01 & {--} \\
        \code{fp8\_dynamic\_per\_token\_quant} & mem & 0.3 & 0.8 & 20 & 70.0 & 0.25 & 0.16 \\
        \code{linear\_sigmoid\_mul} & comp & 1018 & 0.7 & 9 & 57.4 & 0.52 & 0.45 \\
        \code{per\_token\_group\_quant\_fp8} & mem & 0.3 & 0.5 & 19 & 44.7 & 0.39 & 0.15 \\
        \code{gated\_rms\_norm} & mem & 1.3 & 0.4 & 7 & 100.0 & 0.34 & 0.29 \\
        \code{l2\_norm} & mem & 0.8 & 0.4 & 13 & 100.0 & 0.44 & 0.41 \\
        \code{moe\_count\_and\_sort} & idx & {--} &0.4 & 5 & 86.7 & 0.00 & 0.00 \\
        \code{fused\_rmsnorm\_quant} & mem & 0.7 & 0.3 & 11 & 34.8 & 0.03 & 0.13 \\
        \code{moe\_sum\_reduce} & mem & 0.4 & 0.3 & 7 & 100.0 & 0.59 & 0.43 \\
        \code{layer\_norm} & mem & 1.7 & 0.1 & 13 & 70.5 & 0.14 & 0.14 \\
        \bottomrule
    \end{tabular}
\end{table}

\PHM{Where each model breaks.} \tabref{tab:app-errors} partitions the $683$
failing units by model and stage. This breakdown tracks generation quality and, more
usefully, the \emph{kind} of failure. The two weakest backends account for $68\%$ of all failures
and fail mostly at compile time: GLM-5.1 alone contributes $288$, dominated by $152$
candidates that raise before producing an artifact, $61$ compile timeouts, and $19$
that are not even valid \code{nn.Module}s; DeepSeek-V4-Pro adds $175$ failures (mostly compile and
numeric failures, plus $26$ runtime timeouts). The three strongest fail differently. GPT-5.5
fails almost entirely on numeric mismatches ($28$ of $30$ units). Opus~4.7's $46$
failures are unique in the panel: half ($23$) are \emph{correct-but-slow}---kernels
that pass numerics but miss the performance budget---rather than the compile or numeric
breakage that dominates the weaker models. The soft, performance-side failure is thus a
signature of the strongest backend, consistent with the residual-roofline-gap reading
of \secref{sec:models-main}.

\begin{table}[t]
    \caption{Failure taxonomy by model over the $683$ failing units (counts, not
    operator-balanced). Compile: exception before artifact (exc), timeout/OOM (t/o),
    or \code{Model} not \code{nn.Module} (struct). Runtime: numeric mismatch,
    timeout (t/o), exception (exc). Perf: correct but over budget (slow).}
    \label{tab:app-errors}
    \centering
    \small
    \setlength{\tabcolsep}{5pt}
    \renewcommand{\arraystretch}{1.15}
    \begin{tabular}{l S[table-format=3.0] S[table-format=2.0] S[table-format=2.0] S[table-format=3.0] S[table-format=2.0] S[table-format=1.0] S[table-format=2.0] S[table-format=3.0]}
        \toprule
        & \multicolumn{3}{c}{Compile} & \multicolumn{3}{c}{Runtime} & {Perf} & \\
        \cmidrule(lr){2-4}\cmidrule(lr){5-7}
        {Model} & {exc} & {t/o} & {struct} & {numeric} & {t/o} & {exc} & {slow} & {Total} \\
        \midrule
        Claude Opus 4.7 & 1 & 0 & 0 & 20 & 0 & 2 & 23 & 46 \\
        GPT-5.5 & 0 & 0 & 0 & 28 & 0 & 2 & 0 & 30 \\
        Qwen3.7-Max & 18 & 0 & 0 & 48 & 0 & 2 & 0 & 68 \\
        Kimi-K2.6 & 20 & 21 & 0 & 33 & 0 & 2 & 0 & 76 \\
        GLM-5.1 & 152 & 61 & 19 & 55 & 0 & 1 & 0 & 288 \\
        DeepSeek-V4-Pro & 70 & 3 & 0 & 73 & 26 & 2 & 1 & 175 \\
        \midrule
        Total & 261 & 85 & 19 & 257 & 26 & 11 & 24 & 683 \\
        \bottomrule
    \end{tabular}
\end{table}

% \newpage
% \input{checklist.tex}

\end{document}